%% file: main.tex
\documentclass[10pt,twocolumn,letterpaper]{article}

\usepackage[pagenumbers]{iccv} %

\input{preamble}
\definecolor{iccvblue}{rgb}{0.21,0.49,0.74}
\usepackage[pagebackref,breaklinks,colorlinks,allcolors=iccvblue]{hyperref}
\usepackage{url}
\usepackage{multirow}
\usepackage{stfloats}
\usepackage[dvipsnames]{xcolor}
\usepackage{soul}
\usepackage{subcaption}

\def\paperID{102} %
\def\confName{ICCV}
\def\confYear{2025}
\def\methodName{CLoRA}

\title{Contrastive Test-Time Composition of Multiple LoRA Models for Image Generation}

\author{Tuna Han Salih Meral$^{1,}\footnotemark[1]$ \quad Enis Simsar$^{2,}$\footnotemark[1] \quad Federico Tombari$^{3,4}$ \quad Pinar Yanardag$^{1}$ \\[2mm]
$^{1}$Virginia Tech \qquad  $^{2}$ETH Zürich \qquad $^{3}$TUM \qquad $^{4}$Google\\
{\tt\small \href{https://clora-diffusion.github.io}{https://clora-diffusion.github.io}}
}   

\begin{document}

\input{sec/teaser}

{
    \renewcommand{\thefootnote}{\fnsymbol{footnote}}
   \footnotetext[1]{Joint first-authors.}
}

\maketitle
\input{sec/0_abstract}    
\input{sec/1_intro}

\input{sec/2_related_work}

\input{sec/3_methodology}

\input{sec/4_experiments}

\input{sec/6_conclusion}

\newpage
{
    \small
    \bibliographystyle{ieeenat_fullname}
    \bibliography{main}
}

\newpage
\onecolumn

\twocolumn
\input{sec/X_appendix}

\end{document}

%% file: sec/teaser.tex
\twocolumn[{
\maketitle
\begin{center}
    \captionsetup{type=figure}
    \vspace{-1em}
\newcommand{\imwidth}{1\textwidth}

\begin{tabular}{@{}c@{}}

\parbox{\imwidth}{\centering \includegraphics[width=0.9\imwidth]{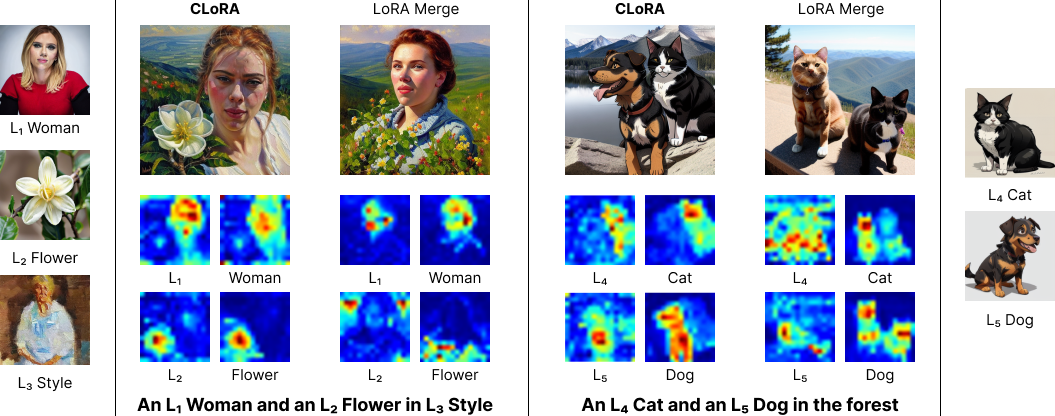}}
\\

\vspace{1em}
\end{tabular}
    \vspace{-2.5em}
    \captionof{figure}{\methodName{} is a training-free method that works on test-time, and uses contrastive learning to compose multiple concept and style LoRAs simultaneously. Using pre-trained LoRA models, such as $L_1$ for a person, and $L_2$ for a specific type of flower, the goal is to create an image that accurately represents both concepts described by their respective LoRAs. Directly combining these LoRA models to compose the image often leads to poor outcomes (see {LoRA Merge}). This failure primarily arises because the attention mechanism fails to create coherent attention maps for subjects and their corresponding attributes. \methodName{} revises the attention maps in test-time to clearly separate the attentions associated with distinct concept LoRAs.} 
    \label{fig:teaser}
\end{center}
}]

%% file: sec/0_abstract.tex
\begin{abstract}

Low-Rank Adaptation (LoRA) has emerged as a powerful and popular technique for personalization, enabling efficient adaptation of pre-trained image generation models for specific tasks without comprehensive retraining. While employing individual pre-trained LoRA models excels at representing single concepts, such as those representing a specific dog or a cat, utilizing multiple LoRA models to capture a variety of concepts in a single image still poses a significant challenge. Existing methods often fall short, primarily because the attention mechanisms within different LoRA models overlap, leading to scenarios where one concept may be completely ignored (e.g., omitting the dog) or where concepts are incorrectly combined (e.g., producing an image of two cats instead of one cat and one dog). We introduce \methodName{}, a training-free approach that addresses these limitations by updating the attention maps of multiple LoRA models at test-time, and leveraging the attention maps to create semantic masks for fusing latent representations. This enables the generation of composite images that accurately reflect the characteristics of each LoRA.  Our comprehensive qualitative and quantitative evaluations demonstrate that \methodName{} significantly outperforms existing methods in multi-concept image generation using LoRAs. %
\end{abstract}

%% file: sec/1_intro.tex
\section{Introduction}
\label{sec:intro}

Diffusion text-to-image models~\cite{ddpm} have revolutionized the generation of images from textual prompts, as evidenced by significant developments in models such as Stable Diffusion~\cite{rombach2022high}, Imagen~\cite{imagen}, and DALL-E 2~\cite{dalle2}. Their applications extend beyond image creation, including tasks like image editing~\cite{blended,blended_latent,diffedit,p2p}, inpainting~\cite{lugmayr2022repaint}, and object detection~\cite{chen2023diffusiondet}. %
As generative models gaining popularity, personalized image generation plays a crucial role in creating high-quality, diverse images tailored to user preferences. Low-Rank Adaptation ~\cite{hu2021lora}, initially introduced for LLMs, has emerged as a powerful technique for model personalization in image generation.   LoRA models can efficiently fine-tune pre-trained diffusion models without the need for extensive retraining or significant computational resources. They are designed to optimize low-rank, factorized weight matrices specifically for the attention layers and are typically used in conjunction with personalization methods like DreamBooth~\cite{ruiz2023dreambooth}.  Since their introduction, LoRA models have gained significant popularity among researchers, developers, and artists~\cite{gandikota2023concept, guo2023animatediff}. For example, Civit.ai\footnote{\url{http://civit.ai}}, a widely used platform for sharing pre-trained models, hosts more than 100K LoRA models~\cite{luo2024stylus} tailored to specific characters, clothing styles, or other visual elements, allowing users to personalize their image creation experiences. %

While existing LoRA models function as plug-and-play adapters for pre-trained models, integrating multiple LoRAs to facilitate the joint composition of  concepts is an increasingly popular task.  The ability to blend a diverse set of elements, such as various artistic styles or the incorporation of unique objects and people, into a cohesive visual narrative is crucial for leveraging compositionality~\cite{huang2023composer,zhang2023adding}. For example, consider a scenario where a user has two pre-trained LoRA models, representing a cat and a dog in a specific style (see Fig. \ref{fig:teaser}). The objective might be to use these models to generate images of this particular cat and dog against various backgrounds or in different scenarios. However, using multiple LoRA models to create new, composite images has proven to be challenging, often leading to unsatisfactory results (see Fig. \ref{fig:teaser}). 

Prior works on combining LoRA models, such as the application of weighted linear combination of multiple LoRAs \cite{ryu2023low}, often lead to unsatisfactory outcomes where one of the LoRA concepts is often ignored. Other approaches~\cite{shah2023ziplora, huang2023lorahub}  train coefficient matrices to merge multiple LoRA models into a new one.  However, these methods are limited by the capacity to merge only a single content and style LoRA~\cite{shah2023ziplora} or by performance issues that destabilize the merging process as the number of LoRAs increases~\cite{huang2023lorahub}. Other methods, such as Mix-of-Show~\cite{gu2023mix}, necessitate training specific LoRA variants such as Embedding-Decomposed LoRAs (EDLoRAs),  diverging from the traditional LoRA models (e.g., civit.ai) commonly used within the community. They also depend on controls like regions defined by ControlNet \cite{zhang2023adding} conditions, which restrict their capability for condition-free generation. More recent works, such as OMG \cite{kong2024omg} utilizes off-the-shelf segmentation methods to isolate subjects during generation, with the overall effectiveness significantly dependent on the accuracy of the underlying segmentation model.

Contrary to these methods, we propose a solution that composes multiple LoRAs at test-time, without the need for training new models or specifying controls. Our approach involves adjusting the attention maps through latent updates during test-time to guide the appropriate LoRA model to the correct area of the image while keeping LoRA weights intact. The significant issues of `attention overlap' and `attribute binding', previously noted in image generation \cite{chefer2023attend, agarwal2023star}, also exist in LoRA models. Attention overlap occurs when specialized LoRA models redundantly focus on similar features or areas within an image. This situation can lead to a dominance issue, where one LoRA model might overpower the contributions of others, skewing the generation process towards its specific attributes or style at the expense of a balanced representation (see Fig. \ref{fig:teaser}). Another related issue is attribute binding, especially occurs in scenarios involving multiple content-specific LoRAs where features intended to represent different subjects blend indistinctly, failing to maintain the integrity and recognizability of each concept. For instance, consider the prompt `An $L_4$ cat and an $L_5$ dog in the forest' in Fig. \ref{fig:teaser}, which depicts two LoRA models tailored for a specific cat and a dog, respectively. The straightforward approach of composing these LoRA models by merging the LoRA weights (see Fig. \ref{fig:teaser} - {LoRA Merge}) struggles to produce the intended results. This is because the $L_4$ attention, which should focus on the cat, blended with the $L_5$ attention, designated for the dog. Therefore, the output incorrectly features two cats, entirely omitting the dog. In contrast, our approach refines the attention maps of the LoRA models in test-time to concentrate on the intended attributes, and produces an image that accurately places both LoRA models in their correct positions (see Fig. \ref{fig:teaser}). 
\methodName{} successfully composes multiple LoRA models while addressing attention overlap and attribute binding, with the following key contributions: %

\begin{itemize}
    \item We present a novel approach based on a  contrastive objective to seamlessly integrate multiple content and style LoRAs simultaneously. Our method dynamically updates latents based on attention maps at test-time and fuses multiple latents using masks derived from cross-attention maps corresponding to distinct LoRA models. Our approach  does not require training and does not require any additional controls unlike common approaches. %

    \item Unlike some of the previous methods, our approach does not need specialized LoRA variants and can directly use community LoRAs on civit.ai in a plug-and-play manner.

    \item We present a comprehensive qualitative and quantitative evaluation of CLoRA's performance, highlighting its superiority over existing approaches for multi-concept image generation with LoRAs. Moreover, our method is highly efficient in both memory usage and runtime, capable of scaling to handle multiple LoRA models. %
    Additionally, we publicly share our source code (see Appendix).

\end{itemize}

%% file: sec/2_related_work.tex
\section{Related Work}
\label{sec:related_work}

\noindent\textbf{Attention-based Methods for Improved Fidelity} 
Text-to-image diffusion models often struggle with fidelity to input prompts, particularly when dealing with complex prompts containing multiple concepts \cite{tang2022daam}. %
Recent advancements in high-fidelity text-to-image diffusion models ~\cite{chefer2023attend, li2023divide, agarwal2023star} share our approach of utilizing attention maps to enhance image generation fidelity. A-Star~\cite{agarwal2023star} and DenseDiffusion~\cite{kim2023dense} refine attention during the image generation process.~\cite{chefer2023attend} address neglected tokens in prompts, while~\cite{li2023divide} propose separate objective functions for missing objects and incorrect attribute binding issues. \cite{xie2023boxdiff} {and} \cite{phung2024grounded} {utilize bounding boxes additional constraint to limit the generation of multiple subjects in constrained areas.}  While these methods tackle attention overlap and attribute binding within a single diffusion model, our approach uniquely addresses these issues across multiple LoRA models. \cite{meral2023conform} use a contrastive approach on a single diffusion model, whereas our technique resolves these challenges across multiple diffusion models (LoRAs), each fine-tuned for distinct objects or attributes.

\noindent\textbf{Personalized Image Generation} 
The field of personalized image generation has evolved significantly, building upon a rich history of image-based style transfer~\cite{efros2023image, hertzmann2023image}. Early advancements came through convolutional neural networks~\cite{gatys2016image, huang2017arbitrary, johnson2016perceptual} and GAN-based approaches~\cite{karras2019style, karras2020analyzing, chong2022jojogan, gal2022stylegan, kwon2023one}. More recently, diffusion models~\cite{ddpm, rombach2022high, ddim} have offered superior quality and text control. In the context of large text-to-image diffusion models, personalization techniques have taken various forms. Textual Inversion~\cite{gal2022image} and DreamBooth~\cite{ruiz2023dreambooth} focus on learning specific subject representations. LoRA~\cite{ryu2023low} and StyleDrop~\cite{sohn2023styledrop} optimize for style personalization. Custom Diffusion~\cite{kumari2023multi} attempts multi-concept learning but faces challenges in joint training and style disentanglement. \cite{zhang2024attention} {uses attention calibration to disentangle multiple concepts from a single image and utilizes these concepts to generate personalized images.}

\noindent\textbf{Merging Multiple LoRA Models} 
The combination of LoRAs for simultaneous style and subject control is an emerging area of research, presenting unique challenges and opportunities. Existing approaches have explored various methods, each with its own limitations. Weighted summation, as proposed by~\cite{ryu2023low}, often yields suboptimal results due to its simplicity. \cite{gu2023mix} suggest retraining specific EDLoRA models for each concept, which limits the approach’s applicability to existing community LoRAs. \cite{wu2023mole} propose composing LoRAs through a mixture of experts, but this method requires learnable gating functions that must be trained for each domain.  Test-time LoRA composition methods, such as Multi LoRA Composite and Switch by~\cite{zhong2024multi}, have also been proposed, but these do not operate on attention maps and may produce unsatisfactory results. ZipLoRA~\cite{shah2023ziplora} synthesizes a new LoRA model based on a style and a content LoRA, however their method falls short in handling multiple content LoRAs.   OMG by~\cite{kong2024omg} utilizes off-the-shelf segmentation methods to isolate subjects during generation, with its performance heavily dependent on the multi-object generation fidelity of diffusion models and the accuracy of the underlying segmentation model. \cite{yang2024lora}{ proposes a training-free approach tackling concept confusion by introducing additional injection and isolation constraints using user-provided bounding boxes.} 
Orthogonal Adaptation \cite{po2024orthogonal} reduces interference by separating attributes across LoRAs and merges them via a weighted average. However, it complicates training by modifying fine-tuning and requiring original data.
LoRACLR~\cite{simsar2025loraclr} merges pre-trained LoRA models using a contrastive objective for scalable multi-concept generation with minimal interference, but relies on user-provided control conditions during inference.
Our approach distinguishes itself by directly addressing attention overlap and attribute binding issues in the context of multiple LoRA models. We incorporate test-time generated masks, enhancing the disentanglement of LoRA models and effectively resolving attention map and attribute binding problems. This offers a more comprehensive solution for high-fidelity, multi-concept image generation, bridging the gap between single-model attention refinement and effective LoRA model composition.

%% file: sec/3_methodology.tex
\section{Background}
\label{sec:background}

\noindent\textbf{Diffusion models.} Our method is implemented on the Stable Diffusion 1.5 (SDv1.5) model, a state-of-the-art text-to-image generation framework for LoRA applications. Stable Diffusion operates in the latent space of an autoencoder, comprising an encoder $\mathcal{E}$ and a decoder $\mathcal{D}$. The encoder maps an input image $x$ to a lower-dimensional latent code $z = \mathcal{E}(x)$, while the decoder reconstructs the image from this latent representation, such that $\mathcal{D}(z) \approx x$. The core of Stable Diffusion is a diffusion model~\cite{ddpm} trained within this latent space. The diffusion process gradually adds noise to the original latent code $z_0$, producing $z_t$ at timestep $t$. A UNet-based~\cite{unet} denoiser $\epsilon_{\theta}$ is trained to predict and remove the noise. The training objective is defined as:
\begin{equation}
    \mathcal{L} = \mathbb{E}_{z_t, \epsilon \sim \mathrm{N(0, I)}, c(\mathcal{P}), t } \left[ \Vert \epsilon - \epsilon_{\theta}(z_t, c(\mathcal{P}), t )\Vert ^ {2} \right]
\end{equation}
where $c(\mathcal{P})$ represents the conditional information derived from the text prompt $\mathcal{P}$. Stable Diffusion employs   CLIP~\cite{radford2021learning}  to embed the text prompt into a sequence $c$, then fed into the UNet through cross-attention mechanisms. In these layers, $c$ is linearly projected into keys ($K$) and values ($V$), while the UNet's intermediate representation is projected into queries ($Q$). The attention at time $t$ is then calculated as $A_t = \mathrm{Softmax}(QK^\intercal / \sqrt{d})$.
These attention maps $A_t$ can be reshaped into $\mathbb{R}^{h \times w \times l}$, where $h$ and $w$ are the height and width of the feature map (typically $16\times16$, $32\times32$, or $64\times64$), and $l$ is the text embedding sequence length. Our work utilizes the $16\times16$ attention maps, which  capture the most semantically meaningful information~\cite{p2p}.

\noindent\textbf{LoRA models.} LoRA fine-tunes large models by introducing rank-decomposition matrices while freezing the base layer. In SD fine-tuning, LoRA is applied to cross-attention layers responsible for text and image connection. %
Formally, a LoRA model is represented as a low-rank matrix pair ($W_{\text{out}}$, $W_{\text{in}}$). These matrices capture the adjustments introduced to the $W$ weights of a pre-trained  model ($\theta$). The updated weights during image generation are calculated as $W' = W + W_{in}W_{out}$. The low-rank property ensures that ($W_{\text{out}}$ and $W_{\text{in}}$) have significantly smaller dimensions compared to full-weight matrices, resulting in a drastically reduced file size for the LoRA model, \eg{,} an SDv1.5 model is 3.44GB, while a LoRA model is 15–100MB. %

\noindent\textbf{Contrastive learning} has emerged as a powerful method in representation learning ~\cite{chen2020simple, oord2018representation}. Its core principle is bringing similar data points closer together in a latent embedding space while pushing dissimilar ones apart. Let $x \in \mathcal{X}$ represent an input data point, with $x^+$ denoting a positive pair (both $x$ and $x^+$ share the same label) and $x^-$ denoting a negative pair (where the data points have different labels). The function $f: \mathcal{X} \rightarrow \mathbb{R}^N$ is an encoder that maps an input $x$ to an N-dimensional embedding vector. Various contrastive learning objectives are proposed such as  InfoNCE (also known as NT-Xent)~\cite{oord2018representation} which \methodName{} utilizes. %

\begin{figure}[t]
  \centering
  \includegraphics[width=\linewidth]{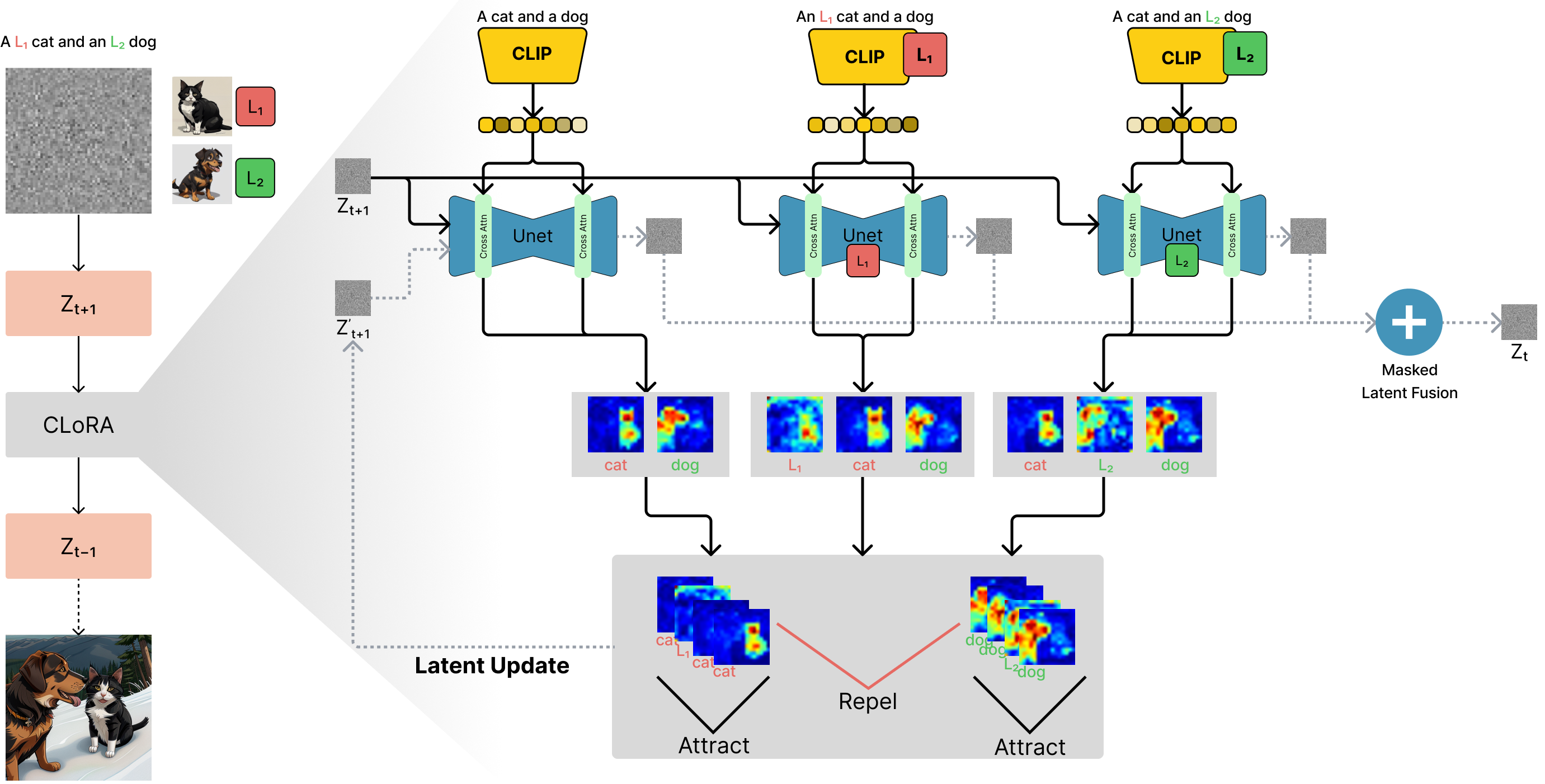}
  \vspace{-1.25em}
  \caption{{Overview of \methodName{}, a training-free, test-time approach for composing multiple LoRA models. Our method accepts a user-provided text prompt, such as `An $L_1$ cat and an $L_2$ dog,' along with their corresponding LoRAs, $L_1$ and $L_2$. \methodName{} applies test-time optimization to attention maps to address attention overlap and attribute binding issues using a contrastive objective.}  %
  }
  \vspace{-1.5em}
  \label{fig:framework}
\end{figure}

\section{Methodology - \methodName{}}
\label{sec:methodology} 

Given a prompt such as `An $L_1$ cat and an $L_2$ dog,' with LoRA models $L_1$ and $L_2$, our method aims to create an image that reflects the text prompt while respecting the corresponding LoRA models (see Fig. \ref{fig:framework}). Our method refines the attention maps of the LoRA models at test-time using a contrastive objective. This objective encourages the attention maps to focus on the intended attributes, thereby resolving issues of attention overlap and attribute binding. Next, we discuss the key components of our contrastive objective and explain how positive and negative pairs are formed.

For simplicity, assume the scenario involves the composition of two LoRA models. {Note that for ease of understanding the positive pairs will be shown in the same color coding such as} \textcolor{BrickRed}{$L_1$ $S_1$} {and} \textcolor{ForestGreen}{$L_2$ $S_2$}. First, we decompose the user-provided prompt into components that align with specific concepts ($S_1$ and $S_2$), defined by different LoRAs ($L_1$ and $L_2$). For example, given the prompt `an \textcolor{BrickRed}{$L_1$ $S_1$} and an \textcolor{ForestGreen}{$L_2$ $S_2$}' (\textit{e.g.,} `An $L_1$ cat and an $L_2$ dog,'), where \textcolor{BrickRed}{$L_1$} and \textcolor{ForestGreen}{$L_2$} represent the personalized concepts for \textcolor{BrickRed}{$S_1$} and \textcolor{ForestGreen}{$S_2$}, respectively, we employ three  prompt variations. First is the original prompt, `an \textcolor{BrickRed}{\textbf{$S_1$}} and an \textcolor{ForestGreen}{\textbf{$S_2$}}'. Second is the $L_1$-applied prompt, `an \textcolor{BrickRed}{$L_1$ \textbf{$S_1$}} and an \textcolor{ForestGreen}{\textbf{$S_2$}}'. Lastly, $L_2$-applied prompt, `an \textcolor{BrickRed}{\textbf{$S_1$}} and an \textcolor{ForestGreen}{$L_2$ \textbf{$S_2$}}'. We then generate corresponding text embeddings using the CLIP model. If the text encoder was fine-tuned during LoRA training, the embeddings are generated using the fine-tuned text encoder. Otherwise, we use the embeddings from the base model. These prompt variations will be used to form positive and negative pairs during the contrastive objective.

During the image generation process, Stable Diffusion utilizes cross-attention maps to guide attention on specific image regions at each diffusion step. However, as discussed before, these attention maps suffer from attention overlap and attribute binding issues, leading to unsatisfactory compositions. We apply a test-time optimization to the attention maps to encourage that each concept (e.g., `\textcolor{BrickRed}{$S_1$}' for the cat or `\textcolor{ForestGreen}{$S_2$}' for the dog) is represented according to their corresponding LoRA. In order to do this, we first categorize cross-attention maps based on their corresponding tokens in the prompts, creating concept groups, \textcolor{BrickRed}{$C_1$} and \textcolor{ForestGreen}{$C_2$}. For the first group, \textcolor{BrickRed}{$C_1$}, we include the cross-attention map for \textcolor{BrickRed}{$S_1$} from the original prompt, cross-attention maps for \textcolor{BrickRed}{$L_1$} and \textcolor{BrickRed}{$S_1$} from the $L_1$-applied prompt, and the cross-attention map for \textcolor{BrickRed}{$S_1$} from the $L_2$-applied prompt. Similarly, for the second group, \textcolor{ForestGreen}{$C_2$}, we include the cross-attention map for \textcolor{ForestGreen}{$S_2$} from the original prompt, the cross-attention map for \textcolor{ForestGreen}{$S_2$} from the $L_1$-applied prompt, and cross-attention maps for \textcolor{ForestGreen}{$L_2$} and \textcolor{ForestGreen}{$S_2$} from the $L_2$-applied prompt. This grouping will be utilized in our contrastive objective to ensure that the diffusion process maintains a coherent understanding of each concept while integrating the stylistic variations introduced by the LoRAs. Separating these concepts will also prevent attention overlap between different concepts, ensuring that each element of the prompt is faithfully represented.

\noindent\textbf{\methodName{} - Contrastive Objective:} We design a contrastive objective during inference to maintain consistency with the input prompt. We used the form of InfoNCE loss due to its fast convergence \cite{oord2018representation}. Our loss function takes pairs of cross-attention maps, processing pairs within the same group as positive and pairs from different groups as negative. For example, given the text prompt `An $L_1$ cat and an $L_2$ dog,' and their corresponding concept groups \textcolor{BrickRed}{$C_1$} (`cat' and $L_1$) and \textcolor{ForestGreen}{$C_2$}  (`dog' and $L_2$), the attention maps of the concept group \textcolor{BrickRed}{$C_1$} form positive pairs. In other words we want the attention map for the cat from the original prompt and the attention map for \textcolor{BrickRed}{$L_1$} from the $L_1$-applied prompt get close to each other since we want $L_1$ LoRA to be aligned with its corresponding subject, cat.  In contrast, the attention maps of different concept groups  \textcolor{BrickRed}{$C_1$} and  \textcolor{ForestGreen}{$C_2$} (e.g., the attention map for cat and dog from the original prompt)  form negative pairs since we want these attention maps to repel each other to avoid attention overlap issue (see Fig. \ref{fig:framework}). The loss function for a single positive pair is expressed as:
\begin{equation}
    \mathcal{L} = -\log \frac{\exp(\mathrm{sim}(A^j, A^{j^+})/\tau)}
    {\sum_{n \in \{j^+, j^-_1, \cdots j^-_N\}} \exp(\mathrm{sim}(A^j, A^n)/\tau)}
    \label{eq:contrastive_loss}
\end{equation}
where cosine similarity $\mathrm{sim}(u, v)$ is defined as $\mathrm{sim}(u, v) = u^T \cdot v / \Vert u \Vert \Vert v \Vert$ where $\tau$ is the temperature, and the denominator includes one positive pair and all negative pairs for $A^j${, $N$ is the number of negative pairs that include $A^j$.} The overall InfoNCE loss is averaged across all positive pairs.

\noindent\textbf{Latent Optimization.} The loss function guides the latents during the diffusion process. The latent representation is updated iteratively similar to~\cite{chefer2023attend} and ~\cite{ agarwal2023star}: $ z^\prime_t = z_t - \alpha_t \nabla_{z_t} \mathcal{L}$ where $\alpha_t$ is the learning rate at step $t$.

\noindent\textbf{Masked Latent Fusion.} In our approach, after a backward step in the diffusion process, we combine the latent representations generated by Stable Diffusion with those derived from additional LoRA models. While the direct combination of these latents is possible as described by~\cite{bar2023multidiffusion}, we introduce a masking mechanism to ensure that each LoRA influences only the relevant regions of the image. This is achieved by leveraging attention maps from the corresponding LoRA outputs to create binary masks. To create the masks, we first extract attention maps for the relevant tokens from each LoRA-applied prompt. For $L_1$, we use the attention maps corresponding to the tokens \textcolor{BrickRed}{$L_1$} and \textcolor{BrickRed}{$S_1$} from the $L_1$-applied prompt, `an \textcolor{BrickRed}{$L_1$ \textbf{$S_1$}} and an \textcolor{ForestGreen}{\textbf{$S_2$}}'. Similarly, for $L_2$, we extract the attention maps for the tokens \textcolor{ForestGreen}{$L_2$} and \textcolor{ForestGreen}{$S_2$} from the $L_2$-applied prompt, `an \textcolor{BrickRed}{\textbf{$S_1$}} and an \textcolor{ForestGreen}{$L_2$ \textbf{$S_2$}}'. Then, we apply a thresholding operation to these attention maps, following a method akin to semantic segmentation described by~\cite{tang2022daam}. For each position $(x, y)$ in the attention map, the binary mask value $M[x,y]$ is determined using the equation $    M[x,y] = \mathbb{I} \left( A[x,y] \geq \lambda \max_{i,j} A[i,j]\right) $
 where $M[x, y]$ represents the binary mask output, $A[x, y]$ is the attention map value at position $(x, y)$ for the corresponding token, $\mathbb{I}(\cdot)$ is the indicator function that outputs 1 if the condition is true (and 0 otherwise), and $\lambda$ is a threshold value between 0 and 1. This thresholding process ensures that only areas with attention values exceeding a certain percentage of the maximum attention value in the map are included in the mask. When multiple tokens contribute to a single LoRA (such as `$L_1$' and `$S_1$' for $L_1$), we perform a union operation on the individual masks to ensure that any region receiving attention from either token is included in the final mask for that LoRA. 
 This masking procedure confines each LoRA's influence to relevant regions, ensuring the generated image retains its integrity while incorporating the stylistic elements specified by the LoRAs. Note that style LoRAs can be applied globally to affect the entire composition or restricted to specific subjects for selective stylization. %

\begin{figure}[t!]
    \centering
    
    \begin{subfigure}[t]{0.95\linewidth}
        \centering
        \includegraphics[width=\linewidth]{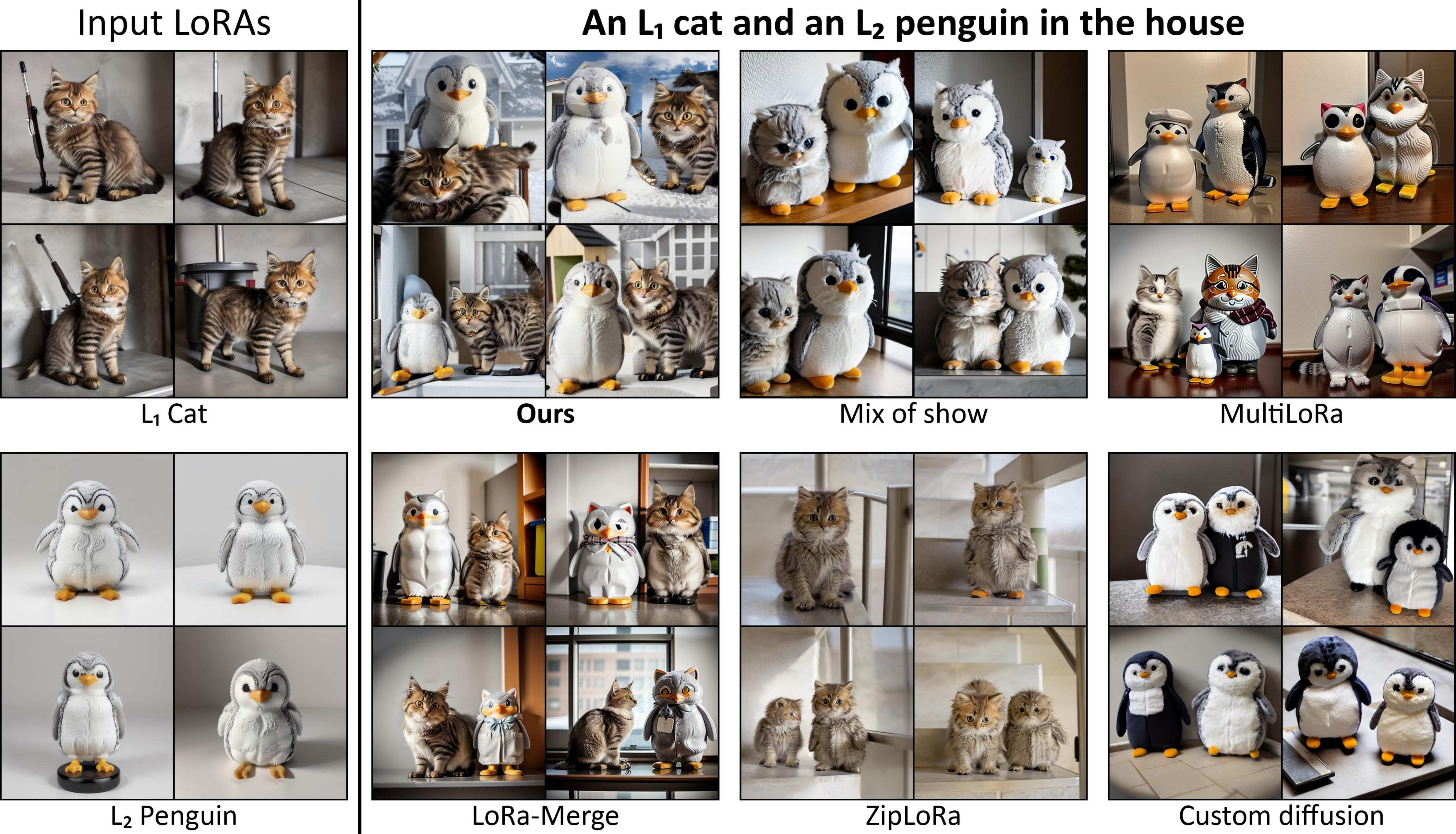}
    \end{subfigure}
    \begin{subfigure}[t]{0.95\linewidth}
        \centering
        \includegraphics[width=\linewidth]{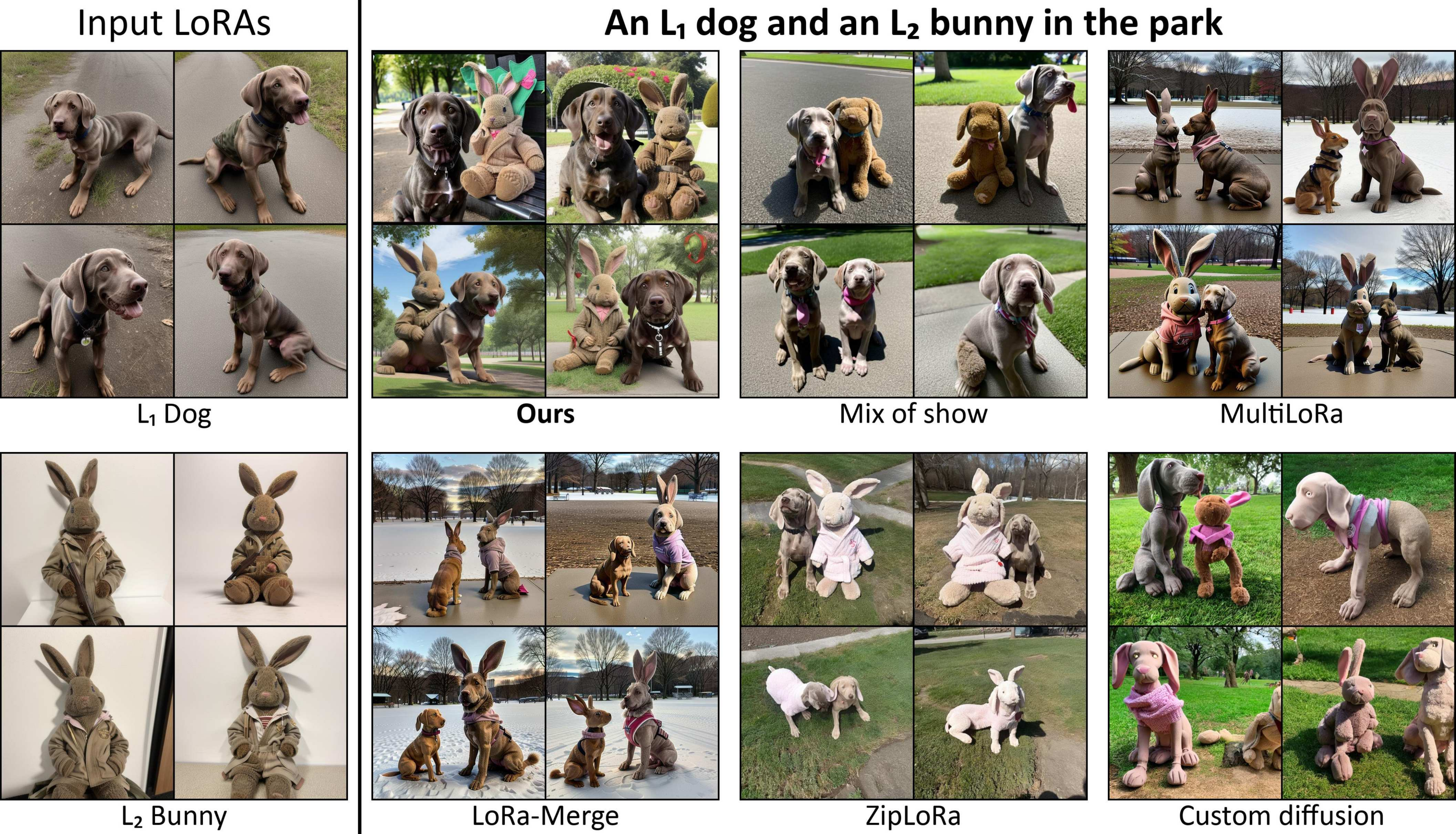}
    \end{subfigure}

    \begin{subfigure}[t]{0.95\linewidth}
      \centering
      \includegraphics[width=\linewidth]{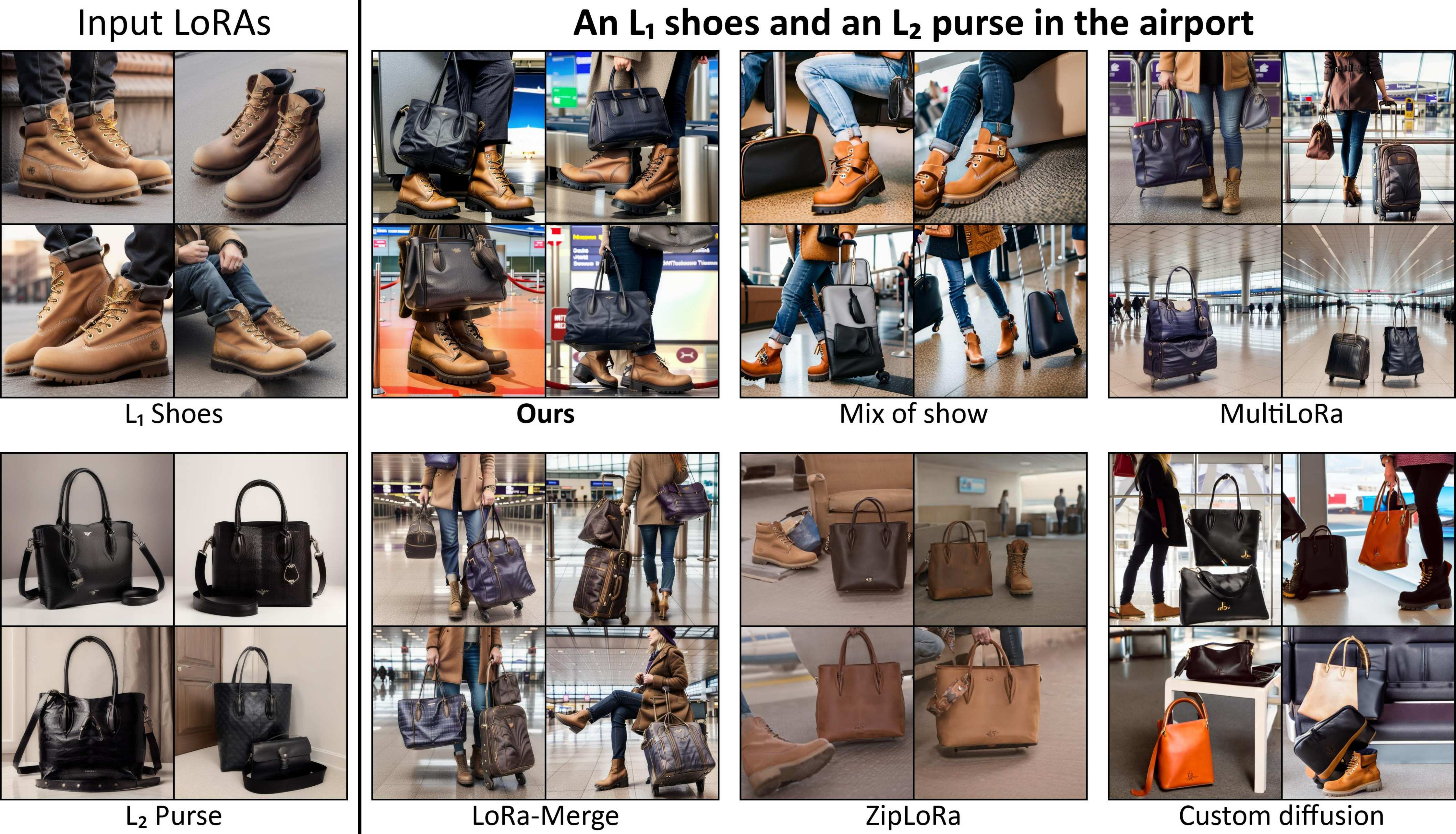}
    \end{subfigure}
  \vspace{-10pt}
  \caption{\textbf{Qualitative Comparison} of \methodName{},  Mix of Show, MultiLoRA, LoRA-Merge, ZipLoRA and  Custom Diffusion.  Our method can generate compositions that faithfully represent the LoRA concepts, whereas other methods often overlook one of the LoRAs and generate a single LoRA concept for both subjects.} %
  \vspace{-1.5em}
  \label{fig:main_comparison} 
\end{figure}

\begin{figure*}[t!]%
  \centering
  \includegraphics[width=.7\textwidth]{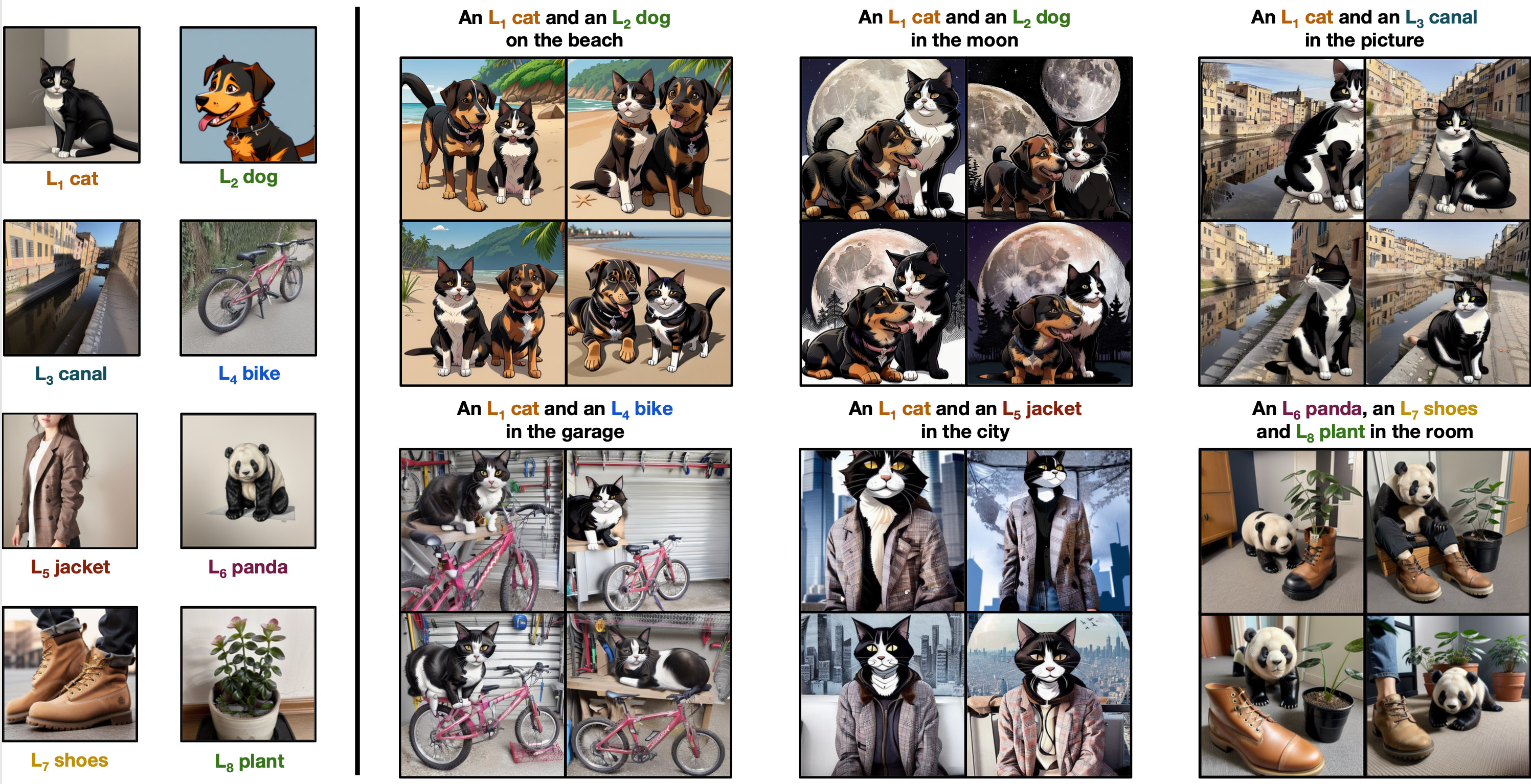}
  \vspace{-0.7em}
  \caption{\textbf{The qualitative results produced by \methodName{}} showcase a range of compositions, including animal-animal, object-object, and animal-object pairs. Left columns display sample images generated by the individual LoRA models. Our approach is successful at composing multiple content LoRAs—for example, combining a \textit{cat} and a \textit{dog}—along with \textit{scene} LoRAs, such as pairing a \textit{cat} with a \textit{canal} scene. Moreover, it demonstrates the capability to integrate more than two LoRAs, exemplified by the composition of a \textit{panda}, \textit{shoe}, and \textit{plant} LoRA (see bottom right).}
  \label{fig:qualitative}
\end{figure*}

\begin{figure*}[ht!]
    \centering
    \begin{subfigure}[t]{0.4\linewidth}
        \centering
        \includegraphics[width=\linewidth]{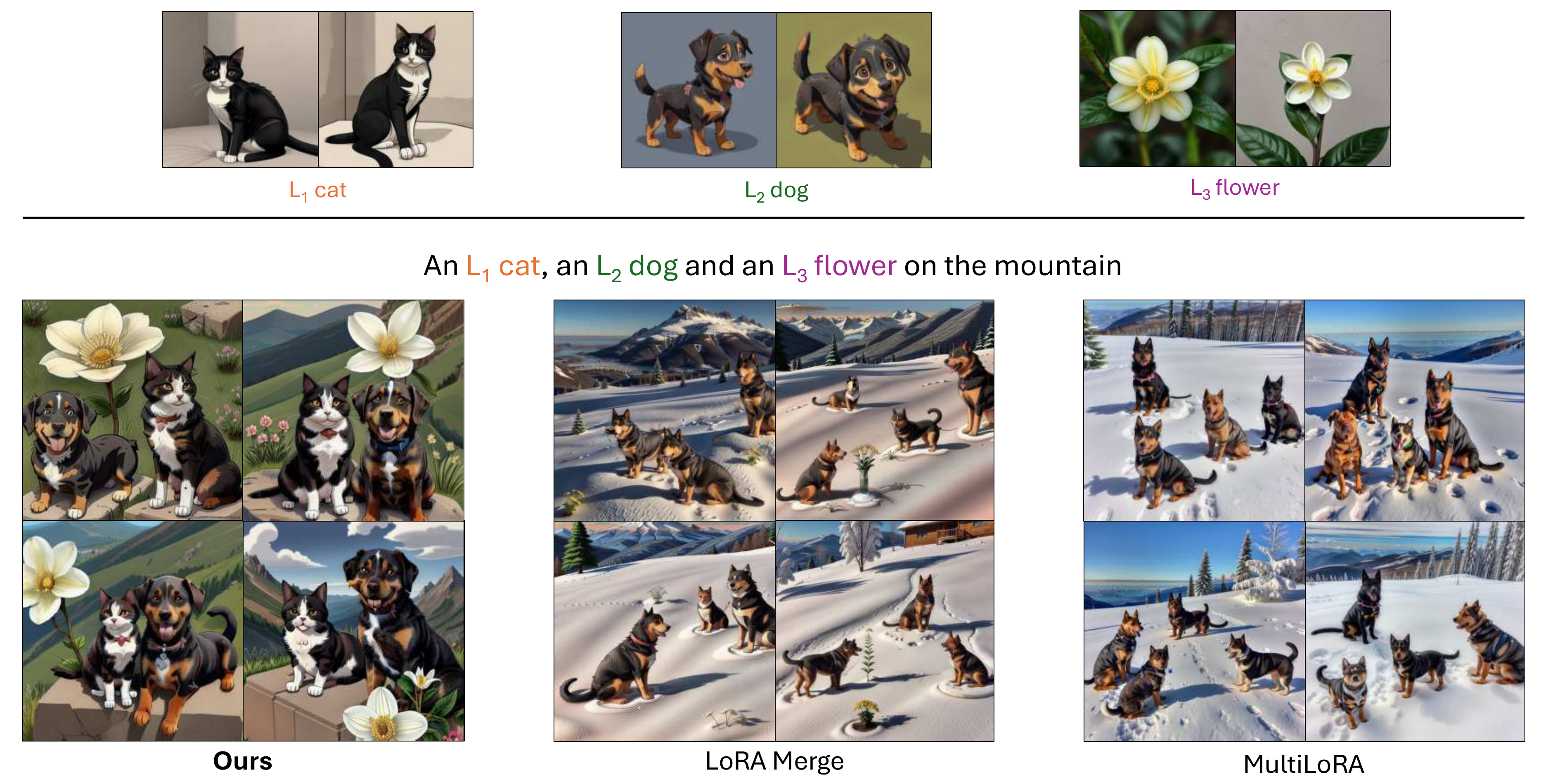}
        \caption{Three LoRA compositions.}
        \label{fig:3loras1}
    \end{subfigure}
    \begin{subfigure}[t]{0.4\linewidth}
        \centering
        \includegraphics[width=\linewidth]{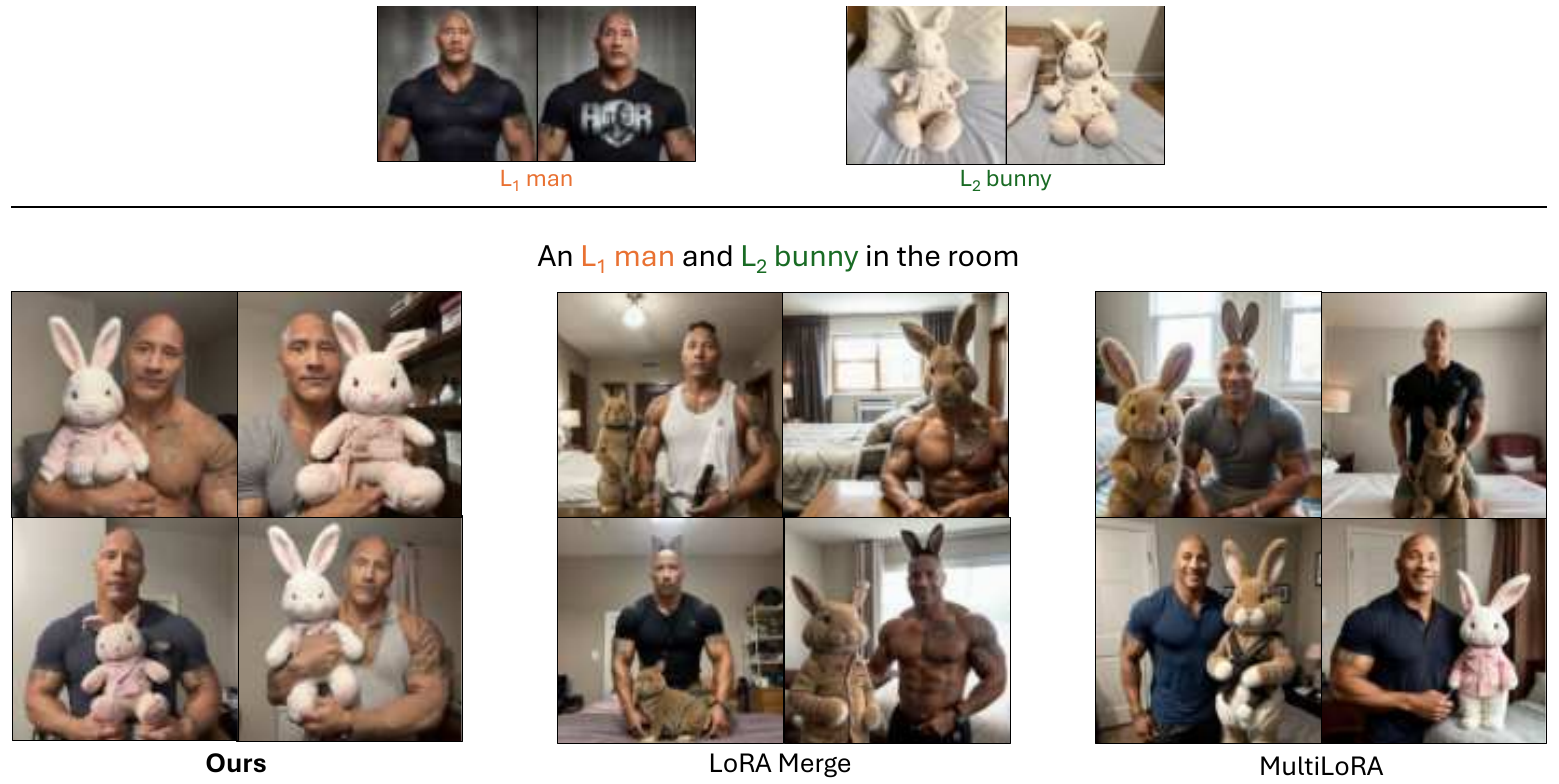}
        \caption{Human subject compositions.}
        \label{fig:realistic}
    \end{subfigure}
    
    \begin{subfigure}[t]{0.7\linewidth}
        \centering
        \includegraphics[width=\linewidth]{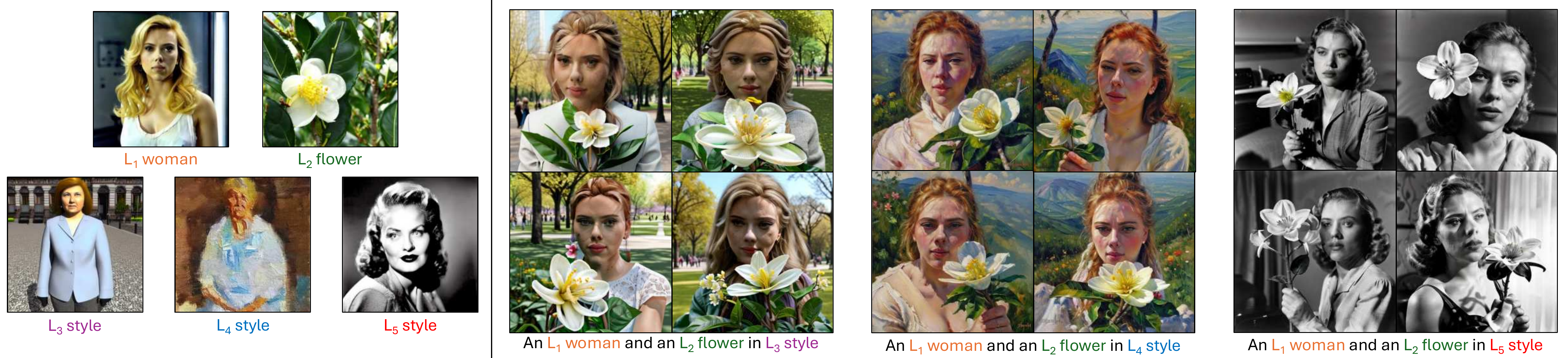}
        \caption{Two subjects and one style compositions.}
        \label{fig:style}
    \end{subfigure}
    \vspace{-10pt}
    \caption{\textbf{Qualitative Results and Comparisons of \methodName{}.} (a) \methodName{} can successfully compose images using three LoRAs. (b) \methodName{} can handle realistic compositions of humans. (c) \methodName{} can seamlessly compose images using style, object, and human LoRAs.}
    \vspace{-1.5em}
\end{figure*}

\begin{figure}[!htb]
  \centering
  \includegraphics[width=0.85\linewidth]{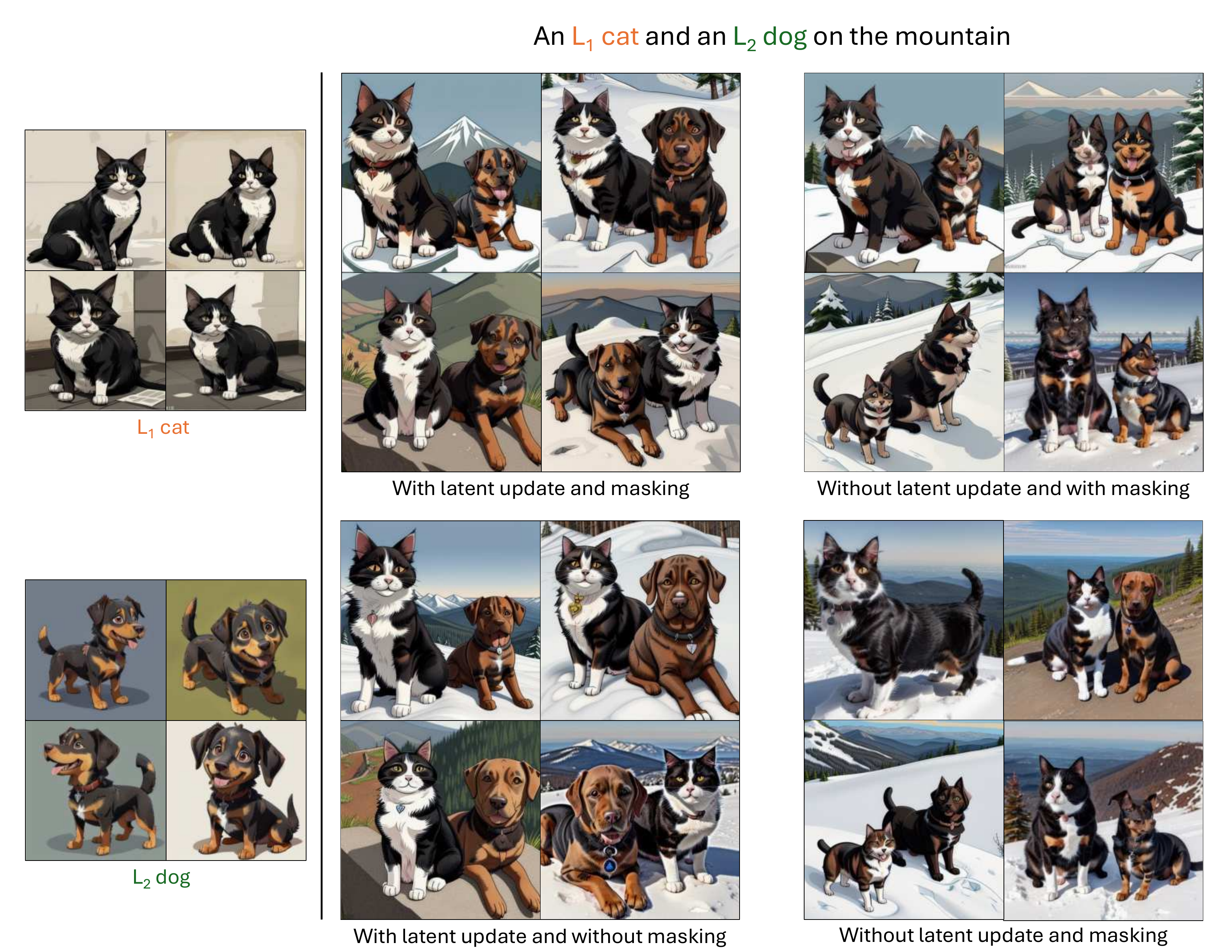}
  \vspace{-.75em}
  \caption{Ablation Study. Using the $L_1$ cat and $L_2$ dog LoRAs, the effects of two key components (latent update and latent masking) can be observed.}
  \label{fig:ablation}
  \vspace{-1.75em}
\end{figure}

%% file: sec/4_experiments.tex
\section{Experiments}
 \label{sec:exp}
In this section, we present qualitative results, along with quantitative comparisons and a user study. %

\noindent\textbf{Datasets.} Due to the absence of standardized benchmarks for composing multiple LoRA models, we compile a set of  131 LoRA models. These models include  custom characters generated with the character sheet trick (see Appendix D) and various concepts from Custom Concept dataset \cite{kumari2023multi}. These models are accompanied by 200 prompts, such as `A plushie bunny and a flower in the forest,' where both `plushie bunny' and `flower' have corresponding LoRA models. See Appendix D for more details. %

\noindent\textbf{Implementation Details.} For each prompt, we use 10 random seeds, running 50 iterations with Stable Diffusion v1.5.  Following \cite{chefer2023attend}, we apply optimization in iterations $i \in \{0, 10, 20\}$, and stop further optimization after $i=25$ to prevent artifacts. For contrastive learning, we set the temperature to $\tau = 0.5$ in Equation~\ref{eq:contrastive_loss}. Image generation was performed on an NVIDIA V100 GPU. {Our approach takes $\approx$ 25 seconds to compose two LoRAs, and can successfully combine up to eight LoRAs on an NVIDIA H100 GPU. %

\noindent\textbf{Baselines.} We compare our results with baselines such as LoRA-Merge \cite{ryu2023low} that merges LoRAs as a weighted combination, ZipLoRA \cite{shah2023ziplora} that synthesizes a new LoRA model based on the provided LoRAs,   Mix-of-Show \cite{gu2023mix} that requires training a specific LoRA type (note that for a fair comparison, our main paper's experiments do not use its additional conditioning), Custom Diffusion \cite{kumari2023multi} and MultiLoRA \cite{zhong2024multi}. For MultiLoRA, we use `Composite', as it outperformed `Switch' \cite{zhong2024multi}. 

\subsection{Qualitative Experiments}

\noindent\textbf{Qualitative Results.}  The qualitative performance of our approach is shown in Fig. \ref{fig:teaser} and \ref{fig:qualitative}. Our method successfully composes images using multiple content LoRAs, such as a \textit{cat} and \textit{dog}, within varied backgrounds like the \textit{mountain} or \textit{moon} (Figs. \ref{fig:teaser} and   \ref{fig:qualitative}). Furthermore, it successfully composes a content LoRA with a scene LoRA, such as  situating the \textit{cat} within a specific \textit{canal} as defined by the scene LoRA (Fig.  \ref{fig:qualitative}). Our method also demonstrates versatility, combining diverse LoRAs, such as a \textit{cat} with a \textit{bicycle} or \textit{clothing} (Fig.  \ref{fig:qualitative}). Notably, it handles compositions involving more than two LoRAs, as illustrated by a \textit{panda}, \textit{shoe}, and \textit{plant} in the bottom right of Fig. \ref{fig:qualitative}.

\noindent\textbf{Qualitative Comparison}  We provide a qualitative comparison between our method and several methods in Fig.~\ref{fig:main_comparison}, focusing on animal-animal and object-object compositions. Each comparison visualizes four randomly generated compositions using our method, Mix-of-Show \cite{gu2023mix}, MultiLoRA \cite{zhong2024multi}, LoRA-Merge \cite{ryu2023low}, ZipLoRA \cite{shah2023ziplora}, and Custom Diffusion \cite{kumari2023multi}. Our method faithfully captures both concepts from the corresponding LoRA models without attention overlap issues. For example, in a prompt such as “An $L_1$ cat and an $L_2$ penguin in the house” (where $L_1$ represents a cat LoRA and $L_2$ a plush penguin LoRA), Mix-of-Show blends the two objects, producing either two plush penguins while ignoring the \textit{cat} or a single \textit{cat} with plush-like features (Fig.~\ref{fig:main_comparison}, top). MultiLoRA fails to resemble the specific LoRA models, producing either two \textit{cats} or two \textit{penguins}. LoRA-Merge generates a \textit{cat} that somewhat aligns with the intended LoRA but does not capture the \textit{penguin} accurately. ZipLoRA frequently fails to incorporate the plush \textit{penguin}, instead creating two \textit{cats} due to its design constraints for combining multiple content LoRAs. Similarly, Custom Diffusion often overlooks the \textit{cat} LoRA entirely, focusing only on generating the plush \textit{penguin}. Similar observations can be made for Fig.\ref{fig:main_comparison} (bottom). Our method successfully generates both elements within a composition, positioning a specific pair of shoes and a purse as dictated by different LoRA models (Fig.\ref{fig:main_comparison}, bottom). In contrast, other approaches frequently miss one of the elements or create objects that do not match the characteristics outlined by the respective LoRAs. This problem is evident in Fig.~\ref{fig:main_comparison} (middle), where the attributes of the bunny LoRA tend to blend with the dog LoRA, leading to a dog that exhibits the bunny’s plushie features. See Appendix C for additional comparisons.

\noindent\textbf{Composition with three LoRA models.} We evaluate the ability to compose with more than two LoRA models in Fig. \ref{fig:3loras1}. Our method successfully maintains the characteristics of each LoRA in the composite image, while other methods struggle to create coherent compositions, often blending multiple models together\footnote{Some methods were excluded because they could not compose three LoRAs \cite{shah2023ziplora}, or require additional controls \cite{gu2023mix}.}.  Moreover,  Fig. \ref{fig:style} shows sample compositions using 3 LoRAs that corresponds to style, object and human LoRAs.

\noindent\textbf{Composition with human subjects.} We compare the composition of human subjects in Figs. \ref{fig:teaser} and \ref{fig:realistic}. Our method seamlessly composes subjects with objects, preserving the distinct properties of each LoRA. Other methods often struggle to integrate both elements effectively (see Fig. \ref{fig:realistic}).

\noindent\textbf{Composition with style LoRAs.} Our approach can blend both style and concept LoRAs (see Figs. \ref{fig:teaser} and \ref{fig:style}). The results show that our method captures the unique features of each content LoRA (e.g., a flower and a human), while applying the style LoRA consistently across the entire image.

\begin{table*}
  \centering
  \hfill
  \begin{subtable}{0.62\linewidth}
    \centering
        \setlength{\tabcolsep}{7pt}
        \resizebox{\linewidth}{!}{
        \begin{tabular}{c c | c | c c | c  | c | c }
            \toprule
              && LoRA Merge & Composite  & Switch  & ZipLoRA  &  Mix-of-Show   & \textbf{Ours} \\ \midrule
     \multirow{3}{*}{\rotatebox[origin=c]{90}{%
                \begin{tabular}{@{}c@{}}
                    DINO
                \end{tabular}%
            }}  & Min. & 0.376 $\pm$ 0.041 & 0.288 $\pm$ 0.049& 0.307 $\pm$ 0.055 & 0.369 $\pm$ 0.036 & 0.407 $\pm$ 0.035 & \textbf{0.447 $\pm$ 0.035}\\
     & Avg. & 0.472 $\pm$ 0.036 & 0.379 $\pm$ 0.045 & 0.395 $\pm$ 0.053 & 0.496 $\pm$ 0.030 & 0.526 $\pm$ 0.024&\textbf{0.554 $\pm$ 0.028}\\
                          &Max. &    0.504 $\pm$ 0.038 & 0.417 $\pm$ 0.046 & 0.432 $\pm$ 0.055 & 0.533 $\pm$ 0.032&   0.564 $\pm$ 0.024 & \textbf{0.593 $\pm$ 0.024}\\
            \midrule 
                        \multirow{3}{*}{\rotatebox[origin=c]{90}{%
                \begin{tabular}{@{}c@{}}
                    {CLIP-I}
                \end{tabular}%
            }}  
            & {Min.} & {0.641 $\pm$ 0.029} & {0.614 $\pm$ 0.035} & {0.619 $\pm$ 0.039} & {0.659 $\pm$ 0.022} & {0.664 $\pm$ 0.023} & {\textbf{0.683 $\pm$ 0.017}} \\
            & {Avg.} & {0.683 $\pm$ 0.029} & {0.654 $\pm$ 0.035} & {0.659 $\pm$ 0.036} & {0.707 $\pm$ 0.021} & {0.712 $\pm$ 0.022} & {\textbf{0.725 $\pm$ 0.017}} \\
            & {Max.} & {0.714 $\pm$ 0.028} & {0.690 $\pm$ 0.033} & {0.695 $\pm$ 0.036} & {0.740 $\pm$ 0.021} & {0.744 $\pm$ 0.023} & {\textbf{0.756 $\pm$ 0.017}} \\ \midrule
            \multicolumn{2}{c|}{{CLIP-T}} & {0.814 $\pm$ 0.054} & {0.833 $\pm$ 0.091} & {0.822 $\pm$ 0.089} & {0.767 $\pm$ 0.081} & {0.760 $\pm$ 0.074} & {\textbf{0.862 $\pm$ 0.052}} \\ \midrule
            \multicolumn{2}{c|}{User Study} & 2.0 $\pm$ 1.10 & 2.11 $\pm$ 1.12 & 1.98 $\pm$ 1.14 & 2.81 $\pm$ 1.18 & 2.03 $\pm$ 1.12 & \textbf{3.32 $\pm$ 1.13}\ \\ \bottomrule
        \end{tabular}}
        \caption{Average, Minimum/Maximum DINO image-image similarities, and {CLIP-I and CLIP-T metrics} between the merged prompts and individual LoRA models, User Study. For all metrics, the higher, the better. \label{table:results}} %
  \end{subtable}
  \hfill
  \begin{subtable}{0.35\linewidth}
    \centering
    \resizebox{\linewidth}{!}{
        \begin{tabular}{l|ccc}
        \toprule
        \textbf{Method}              & \textbf{CivitAI} & \textbf{VRAM} & \textbf{Runtime} \\  \midrule
        Custom Diffusion             & \(\times\)                     & 28GB + 8GB                           & 4.2 min + 3.5s                          \\ 
        LoRA Merge                   & \(\checkmark\)                 & 7GB                                  & 3.2s                                    \\ 
        Composite                    & \(\checkmark\)                 & 7GB                                  & 3.4s                                    \\ 
        Switch                       & \(\checkmark\)                 & 7GB                                  & 4.8s                                    \\ 
        Mix-of-Show                  & \(\times\)                     & 10GB + 10GB                          & 10min + 3.3s                            \\ 
        ZipLoRA                      & \(\checkmark\)                 & 39GB + 17GB                          & 8min + 4.2s                             \\ 
        OMG                          & \(\checkmark\)                 & 30GB                                 & 62s                                     \\ 
        LoRA-Composer                & \(\times\)                     & 51GB                                 & 35s                                     \\ 
        Ours                         & \(\checkmark\)                 & 25GB                                 & 24s                                     \\  \bottomrule
        \end{tabular}
    }
    \caption{{Comparison of methods in terms of CivitAI compatibility, VRAM usage, and runtime (Finetuning and/or Inference).}}\label{tab:method_comparison}

    \end{subtable}
    \vspace{-.5em} \caption{Quantitative Evaluation and Runtime Analysis.} \vspace{-2em}
  \hfill
\end{table*}

\subsection{Quantitative Experiments}
\label{sec:quantitative}
\noindent\textbf{Quantitative Comparison.} We leverage DINO and CLIP \cite{radford2021learning} to assess the quality of images generated by our method and compare methods that combine multiple LoRAs. DINO offers a hierarchical representation of image content, enabling a more detailed analysis of how each LoRA contributes to specific aspects of the merged image. To calculate DINO-based metrics, we first generate separate outputs using each individual LoRA based on the prompt sub-components (e.g., $L_1$ cat' and $L_2$ flower'). Then, we extract DINO features for the merged image and each single LoRA output. Finally, we calculate cosine similarity between the DINO features of the merged image and the corresponding features from each single LoRA output.
    
We utilize three DINO-based metrics: \textit{Average DINO Similarity}, which reflects the overall alignment between the merged image and individual LoRAs averaged across all LoRAs; \textit{Minimum DINO Similarity}, which uses the cosine similarity between the DINO features of the merged image and the least similar LoRA reference output; and \textit{Maximum DINO Similarity}, which identifies the LoRA reference image whose influence is most represented in the merged image. %
For each LoRA model and composition prompts, 50 reference images are generated and DINO similarities are calculated. Prompts used in benchmarks consist of two subjects and a background, such as `an $L_1$ cat and an $L_2$ penguin in the house' (see Fig. \ref{fig:main_comparison}).  The results (see Tab. \ref{table:results}) demonstrate that our method surpasses the baselines in terms of faithfully merging content from LoRAs. 
{Additionally, we include comparisons using CLIP-I (image-to-image similarity) and CLIP-T (image-to-text similarity) metrics to evaluate the performance of our method against competing approaches (see Tab. 1). The results demonstrate that CloRA consistently outperforms other methods across both metrics, highlighting its  ability to generate images that align with the intended concepts and prompts.}

\noindent\textbf{User Study.}  To further validate \methodName{}, we conducted a user study involving 50 participants recruited through Prolific platform \cite{prolific}. Each participant was shown four generated images per composition from different methods and asked to rate how faithfully each method preserved the concepts represented by the LoRAs (on a scale from 1 = ``Not faithful" to 5 = ``Very faithful"). As presented in Tab.~\ref{table:results}, our method consistently outperforms the baselines, achieving higher scores for faithful representation of concepts.

\noindent\textbf{Ablation Study} Our method integrates two key components to generate compositions with multiple LoRAs: \textit{Latent Update} and \textit{Latent Masking}. \textit{Latent Update} employs our contrastive objective to direct the model's attention precisely towards the concepts specified by each LoRA, preventing misdirection and attention to irrelevant areas. Without this component, the model could erroneously generate duplicate objects or incorrect attribute connections (e.g., producing two dogs instead of a cat and a dog),  as shown in Fig. \ref{fig:ablation}. \textit{Latent Masking} protects the identity of the main subject during generation. Without masking, every pixel would be influenced by all prompts, leading to inconsistencies and loss of identity in the final image.  Together, these components enhance   composition process, enabling users to introduce specific styles or variations into designated regions guided by multiple LoRAs.

\noindent\textbf{Runtime Comparison.} We compare various methods on CivitAI \cite{civitai_website} compatibility, VRAM usage, and runtime in Tab.~\ref{tab:method_comparison}, with evaluations conducted on an NVIDIA H100 GPU (80GB VRAM). Among the  compared methods, only some are compatible with CivitAI. Our method offers a favorable balance of VRAM (25GB) and fast inference, making it efficient for multi-concept image generation.
Figure~\ref{fig:runtime} illustrates how VRAM and inference time scale with the number of LoRAs (e.g., 25 GB and 24s for 2 LoRAs vs. 80 GB and 96s for 8 LoRAs). While increasing LoRAs enhances flexibility, the growth in computational demands remains predictable, underscoring the practicality of our approach for multi-concept applications. Our method scales to compositions with 3-5 LoRAs while maintaining comparable performance to methods that require additional pose or segmentation conditioning (see supplementary for detailed analysis).

\begin{figure}[!ht] %
  \begin{center}
    \includegraphics[width=1\linewidth]{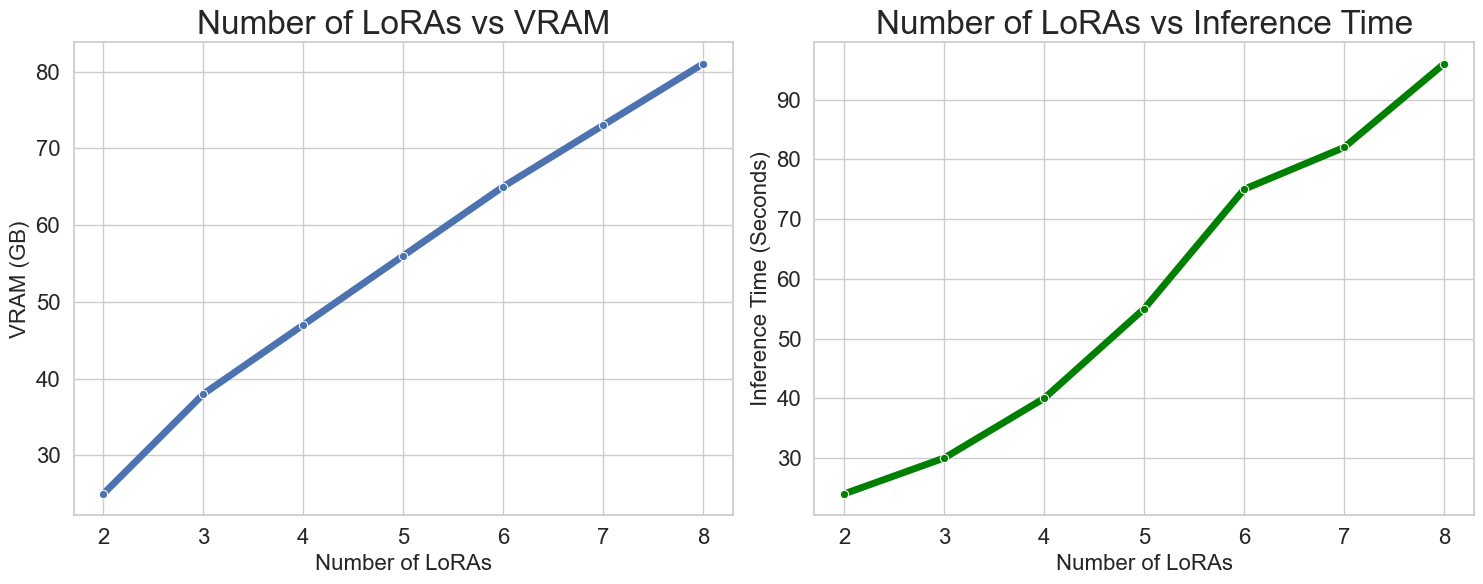}
  \end{center}
  \vspace{-15pt}
  \caption{\textbf{Analysis on Runtime.} {Number of LoRAs vs. VRAM usage, and inference time.}}
  \label{fig:runtime}
  \vspace{-15pt}
\end{figure}

%% file: sec/6_conclusion.tex
\section{Conclusion} We present a training-free method, \methodName{}, for integrating multiple LoRAs to compose images.  Our approach addresses the limitations of existing methods by dynamically adjusting attention maps in test-time, ensuring each LoRA guides the diffusion process toward its designated subject.   Our experimental results demonstrate that \methodName{} significantly outperforms existing baselines across various metrics, including DINO-based similarity, CLIP alignment, and user study evaluations, showcasing its robustness in faithfully representing and blending multiple LoRAs. Unlike other methods, our approach does not require training specific LoRAs and is compatible with a wide range of community-developed LoRAs available on platforms like Civit.ai. By making our source code and LoRA collection public, we aim to promote transparency and reproducibility, as well as encourage further advancements in this area. %

%% file: sec/X_appendix.tex
\setcounter{section}{0}
\renewcommand{\thesection}{\Alph{section}}

\section{Ethics Statement \& Limitations}
\subsection{Ethics Statement}

While our method democratizes creativity by simplifying the process of art creation, it also introduces ethical considerations that must be taken into account. Our method enable the generation of personalized images with minimal effort, and opens the door to transformative opportunities in art and design. However, as noted by \cite{kenthapadi2023generative}, it necessitates a comprehensive and thoughtful discourse around their ethical use to prevent potential abuses. In addition to these concerns, our user study strictly adheres to anonymity protocols to safeguard participant privacy.

The capability of our method to effortlessly generate personalized images also poses risks of misuse in several harmful ways, such as the creation of deepfakes. These can be used to forge identities or manipulate public opinion, a concern  underscored by \citet{korshunov2018deepfakes}.

\input{sec/5_limitations}
\section{User Study Details}

\begin{figure}[!htbp]
   \centering
   \includegraphics[width=\linewidth]{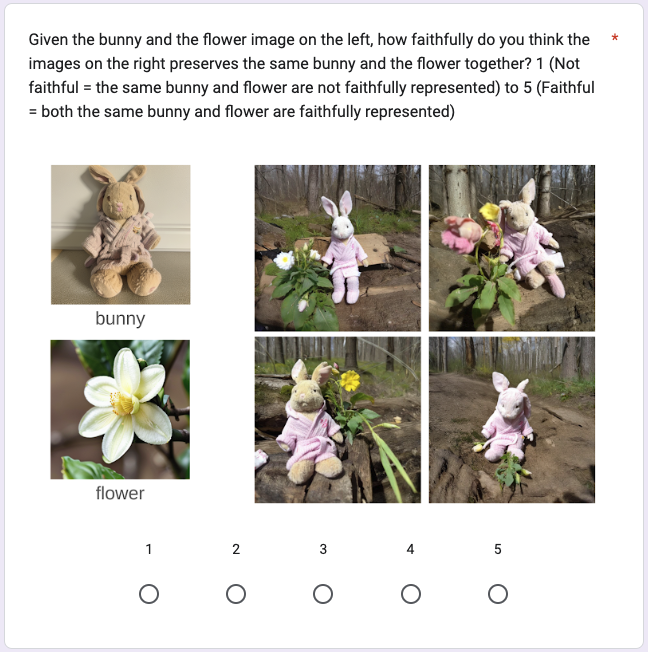}
   \caption{\textbf{Screenshot of our user study.} {Each participant was shown images generated by LoRA models (on the left) and 4 images generated by the method (ours or competitors). Users are then requested to rate from 1-5 (Not faithful/Faithful) based on how well the generated images reflect the concepts depicted in the LoRA models.}}
  \label{fig:userstudy}
 \end{figure}

{We recruited 50 participants through the Prolific platform\footnote{\url{http://prolific.com}.}. Each participant was shown 48 images, and asked to rate how faithfully each method preserved the concepts represented by the LoRAs (on a scale from 1 = ``Not faithful" to 5 = ``Very faithful"). The order of images were randomized per participant. Please see Fig.} \ref{fig:userstudy}  {to see a screenshot of our user study.}

\section{Style LoRA Usage Modes}
\label{appx:style_lora}

\begin{figure}[!htbp]
     \centering
     \includegraphics[width=\linewidth]{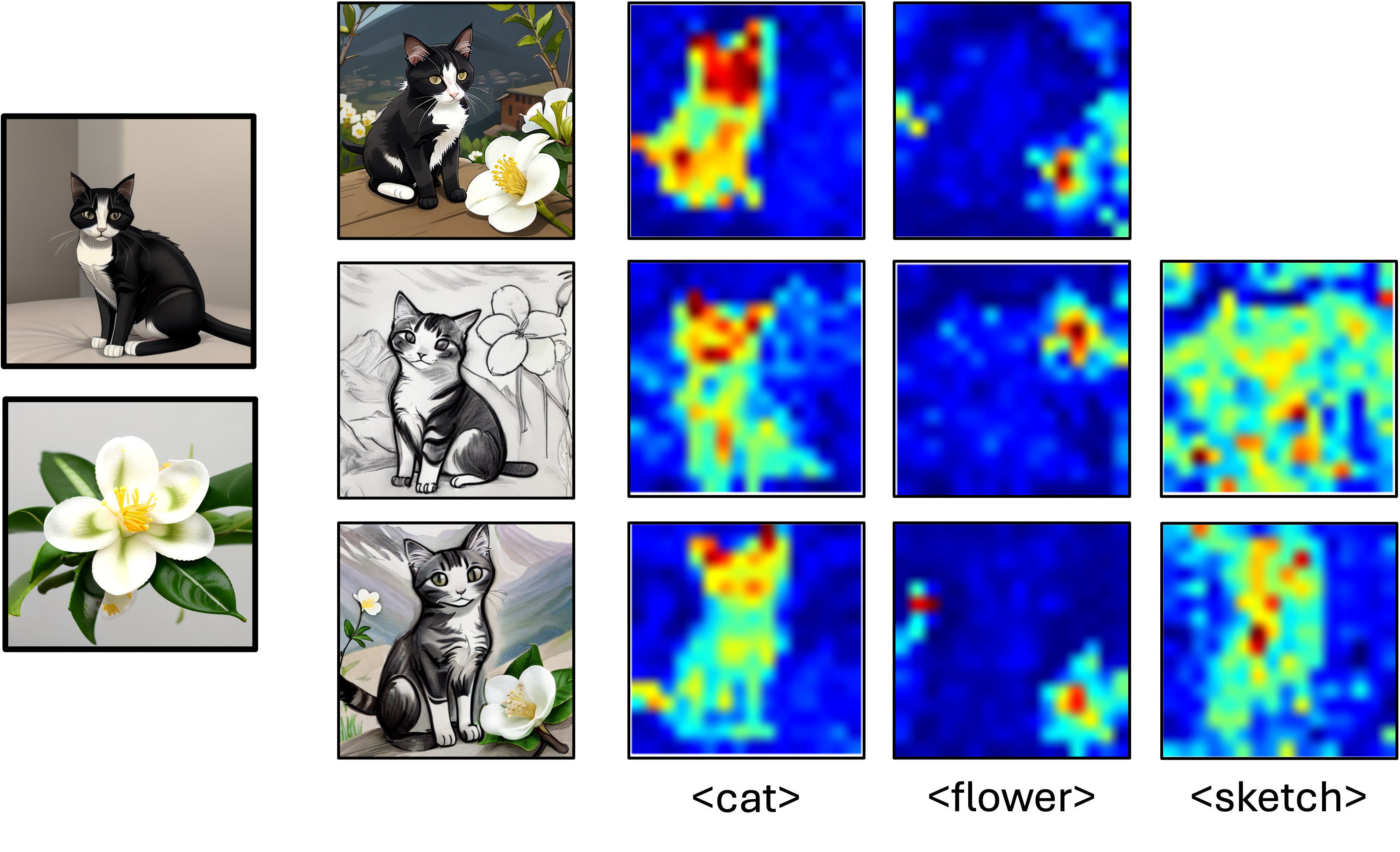}
     \caption{{Attention maps corresponding to subject and style LoRAs using \methodName{}}}
     \label{fig:style-lora-maps}
 \end{figure}

Our framework supports versatile use of style LoRAs: they can be applied \underline{globally} to the complete composition or \underline{restricted} to a single LoRA, according to the user's needs. This flexibility allows users to achieve different artistic effects depending on their creative intent. In \Cref{fig:style-lora-maps}, the top row shows how model attends to each subject token. 

\textbf{Global Style Application:} When a style LoRA is applied globally alongside all prompts, its attention mask spans the whole image and the style affects every pixel, matching standard practice. This mode is useful when users want a consistent artistic style across the entire composition. In \Cref{fig:style-lora-maps}, middle row shows how model attends to each subject and style LoRA token when style is applied globally. 

\textbf{Subject-Specific Style Application:} When the same style LoRA is activated together with a particular subject LoRA (e.g., pairing a sketch-style LoRA with the cat LoRA in a scene that also contains a flower), the contrastive loss confines the stylistic effect to the cat region and leaves the flower untouched. This approach enables users to apply different styles to different subjects within the same composition. In \Cref{fig:style-lora-maps}, bottom row shows how model attends to each subject and style LoRA token when style is applied to only one subject. 

\section{Additional Results}

\subsection{Multi-Class Object Handling}
\label{appx:multi_class}

Our method is capable of handling prompts containing multiple objects from the same super-class, such as multiple people or two cats. Since we assign respective LoRAs to individual tokens (e.g., L1 will be a positive pair with the first "person" token, while L2 will be paired with the second "person" token) and our separation mechanism hinges on LoRA-specific attention groups rather than coarse class labels, our method can handle such cases effectively.

This capability is demonstrated in the quantitative comparison on compositions with 3-5 LoRAs containing multiple people, where our method performs comparably to state-of-the-art methods that use special conditions to enforce separation, while our method achieves this without any auxiliary pose or conditioning inputs.

\subsection{Similar Subject Compositions and Complex Scenarios}

\begin{figure*}[!htbp]
     \centering
     \includegraphics[width=\linewidth]{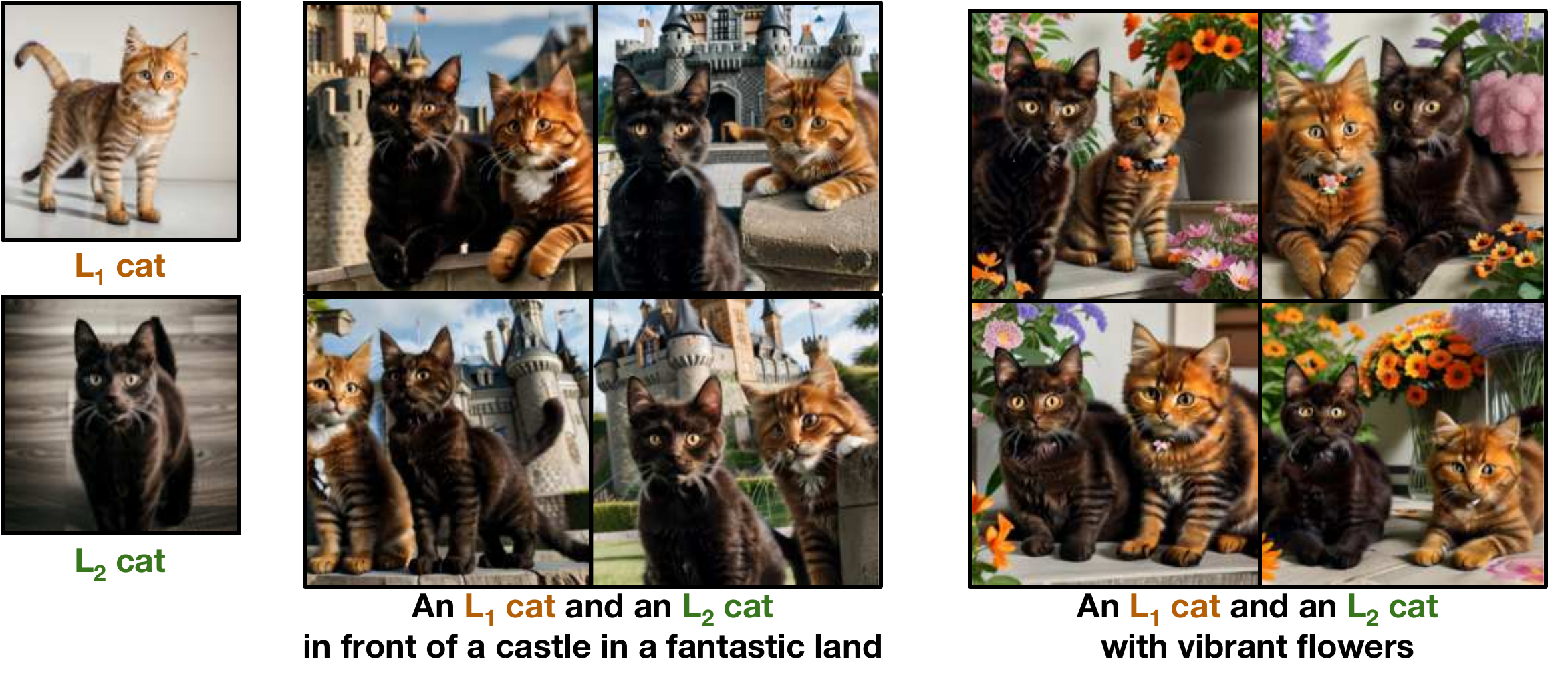}
     \caption{{Qualitative results showing that \methodName{} is capable of generating images using LoRAs that has similar subjects.}}
     \label{fig:similar-subjects}
 \end{figure*}

\begin{figure*}[!htbp]
     \centering
     \includegraphics[width=\linewidth]{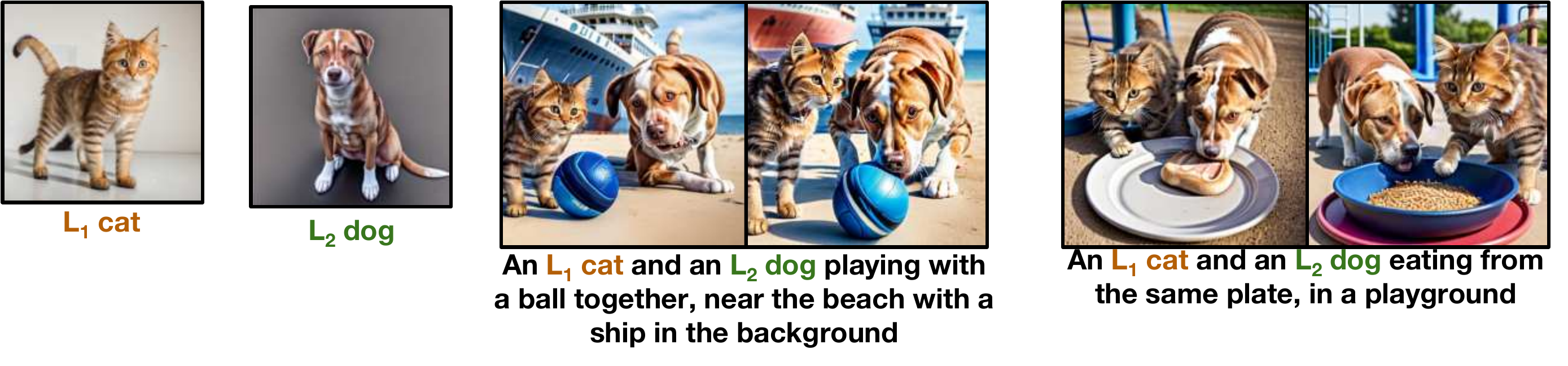}
     \caption{{Qualitative Results showing that \methodName{} is capable of composing images in complex interacting scenes.}}
     \label{fig:complex-interacting}
 \end{figure*}

\Cref{fig:similar-subjects} shows \methodName{}'s capabilities of generating images with similar subjects. \Cref{fig:complex-interacting} showcases the \methodName{}'s ability to merge LoRAs in complex and interacting scenes. Our method can handle visually complex scenes that contain many objects (e.g., bottles, plates, sea on the background, or ship and ball within the same scene). These visuals show that each LoRA subject retains its unique attributes without any cross-subject leakage.

\section{Details of Benchmark LoRA Collection}
 \label{appx:dataset}

{We propose 131 pre-trained LoRA models and 200 text-prompts for multi-LoRA composition. The details of our dataset is given below.}
 
\subsection{Datasets}
This study leverages two key datasets for benchmark:
\begin{itemize}
    \item \textbf{Custom collection:} We generated custom characters such as cartoon style \textit{cat} and \textit{dog}, created using the \textit{character sheet} trick \footnote{\url{https://web.archive.org/web/20231025170948/https://semicolon.dev/midjourney/how-to-make-consistent-characters}} popular within the Stable Diffusion community. This set comprises 20 unique characters,  where we trained a LoRA per character.

    \item \textbf{CustomConcept101:} We used a popular  dataset~\cite{kumari2023multi} CustomConcept101 that includes several diverse objects such as   \textit{plushie bunny}, \textit{flower}, and \textit{chair}. All 101 concepts are utilized. 
\end{itemize}
 
Leveraging the datasets above, we trained LoRAs to  represent each concept, totaling to 131 LoRA models. For every competitor, the base stable diffusion model cited in the relevant paper is used. For instance, ZipLoRA~\cite{shah2023ziplora} employs SDXL, while MixOfShow~\cite{gu2023mix} utilizes EDLoRA alongside SDv1.5. Similarly, our method uses SDv1.5. {Note that while the majority of our concepts are derived from CustomConcept101 dataset, the contribution of our  benchmark LoRA collection is the 131 LoRA models and additional 200 text prompts.}

\subsection{Experimental Prompts}

To evaluate the merging capabilities of the methods, we created 200 text prompts  designed to represent various scenarios such as (the corresponding LoRA models are indicated within paranthesis):  

\begin{itemize}
    \item A cat and a dog in the mountain (blackcat, browndog) 
    \item A cat and a dog at the beach (blackcat, browndog) 
    \item A cat and a dog in the street (blackcat, browndog) 
    \item A cat and a dog in the forest (blackcat, browndog) 
    \item A plushie bunny and a flower in the forest (plushie\_bunny and flower\_1) 
    \item A cat and a flower on the mountain (blackcat, flower\_1)
    \item A cat and a chair in the room  (blackcat, furniture\_1)
    \item A cat watching a garden scene intently from behind a window, eager to explore. (blackcat, scene\_garden)
\item A cat playfully batting at a Pikachu toy on the floor of a child's room. (blackcat, toy\_pikachu1)
\item A cat cautiously approaching a plushie tortoise left on the patio. (blackcat, plushie\_tortoise)
\item A cat curiously inspecting a sculpture in the garden, adding to the scenery. (blackcat, scene\_sculpture1)
\end{itemize}

\section{Comparison with LoRA-Composer}
\begin{figure*}[!htbp]
    \centering
    \includegraphics[width=\linewidth]{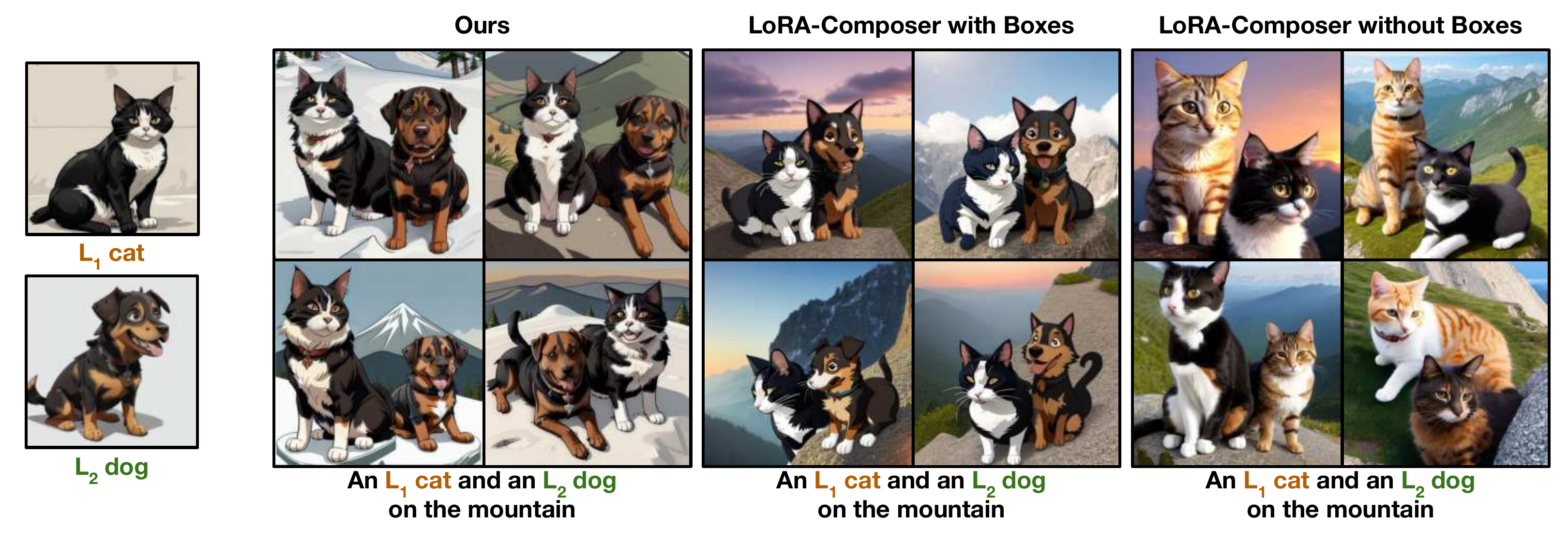}
    \caption{{\textbf{Qualitative comparison with LoRA-Composer.} \methodName{} achieves consistent multi-concept compositions without bounding boxes, unlike LoRA-Composer. Without user-provided bounding boxes, LoRA-Composer method fails to generate the accurate depictions (see rightmost images).}}\label{fig:loracomposer_comparison}
\end{figure*}

{We compare \methodName{} with LoRA-Composer, which operates at test time but requires user-provided bounding boxes, significantly limiting its practicality and ease of use. Additionally, LoRA-Composer is restricted to specific models like ED-LoRA and is incompatible with the wide range of community LoRAs available on platforms like Civit.ai. It also demands substantially more memory, requiring 60GB for generating a composition compared to our method’s 25GB for composing two LoRA models. In contrast, \methodName{} works seamlessly with any standard LoRA models, including community-sourced ones, without relying on bounding boxes or additional conditions. As shown in Fig.} \ref{fig:loracomposer_comparison}, {\methodName{} consistently produces coherent multi-concept compositions, even in challenging scenarios, ensuring broader compatibility and efficiency. For Fig.} \ref{fig:loracomposer_comparison}, {the same seed was used for LoRA-Composer with and without bounding boxes to demonstrate the impact of their presence on the results.}

\section{Multi-LoRA Scalability Analysis}
\label{appx:multi_lora}

We evaluate our method's performance on compositions with 3-5 LoRA models and compare against state-of-the-art multi-LoRA composition approaches: Mix-of-Show~\cite{gu2023mix} and Orthogonal Adaptation. Both baselines rely on ControlNet key-point conditioning and impose extra overhead: Mix-of-Show trains task-specific Ed-LoRAs, while Orthogonal Adaptation merges all input LoRAs into a single adapter. By contrast, our method is entirely test-time: it needs \underline{no key-point cues}, \underline{no specialized LoRAs}, and \underline{no merging or retraining}. 

\begin{figure*}[!htbp]
     \centering
     \includegraphics[width=\linewidth]{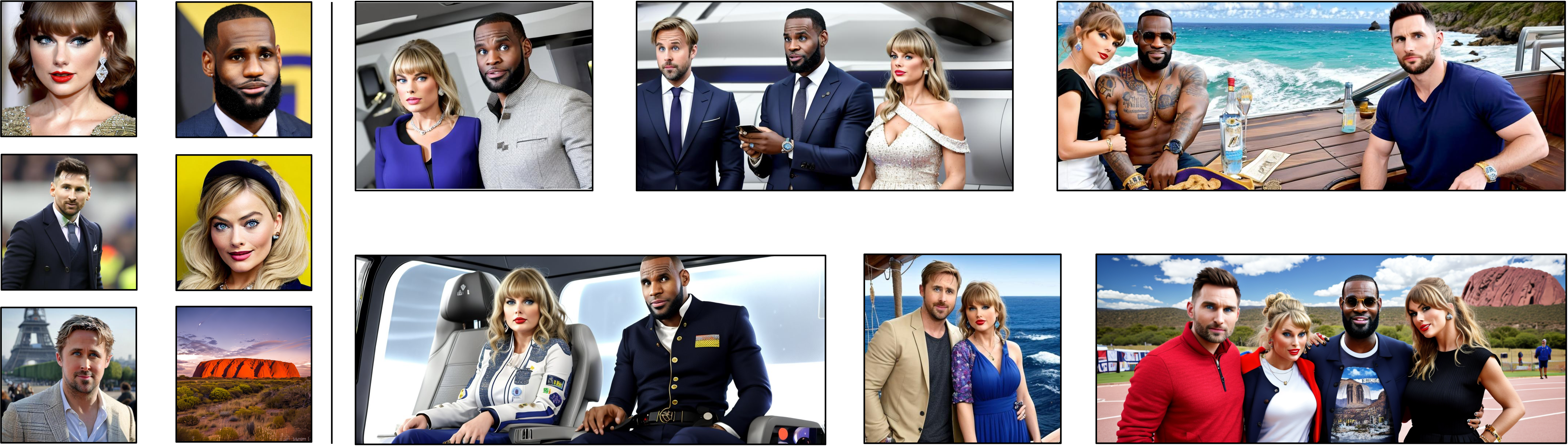}
     \caption{{Various compositions using 2-3-5 Lora models and complex scenes demonstrating \methodName{} is capable of composing multi-subject images in complex scenes.}}
     \label{fig:multi-person}
 \end{figure*}

\Cref{tab:multi_subject} shows quantitative results on 100 compositions with 3-5 input LoRAs. Even without the strong priors such as key-points that explicitly pin down each subject's location, our approach matches the baselines' performance. These results demonstrate that our contrastive test-time strategy maintains high fidelity as the number of subjects and scene complexity increase. \Cref{fig:multi-person} shows qualitative results demonstrating \methodName{}'s capability of generating 2-3-5 people in complex scenes.

\begin{table}[!htb]
    \centering
    \resizebox{\linewidth}{!}{
        \begin{tabular}{llccc}
        \toprule
         &  & Mix-of-Show & Orthogonal Adaptation & \textbf{Ours} \\
        \midrule
        \multirow{3}{*}{\rotatebox[origin=c]{90}{%
            \begin{tabular}{@{}c@{}}
                {CLIP-I}
            \end{tabular}%
        }} & Max.  & 0.688 $\pm$ 0.042 & 0.668 $\pm$ 0.075 & 0.668 $\pm$ 0.065\\
            & Avg. & 0.490 $\pm$ 0.031 & 0.524 $\pm$ 0.042 & 0.525 $\pm$ 0.039\\
            & Min. & 0.371 $\pm$ 0.032 & 0.395 $\pm$ 0.033 & 0.396 $\pm$ 0.034\\
        \midrule
        \multirow{3}{*}{\rotatebox[origin=c]{90}{%
            \begin{tabular}{@{}c@{}}
                {DINO}
            \end{tabular}%
        }} & Max.  & 0.574 $\pm$ 0.078 & 0.548 $\pm$ 0.093 & 0.543 $\pm$ 0.080\\
            & Avg. & 0.351 $\pm$ 0.039 & 0.343 $\pm$ 0.066 & 0.347 $\pm$ 0.054\\
            & Min. & 0.155 $\pm$ 0.046 & 0.158 $\pm$ 0.058 & 0.161 $\pm$ 0.058\\
        \bottomrule
        \end{tabular}
    }
    \caption{Quantitative comparison of Mix-of-Show (MoS), Orthogonal Adaptation (Orth) and our method on 3-, 4-, 5-subject generation using CLIP-I and DINO similarity metrics.}
    \label{tab:multi_subject}
\end{table}

\section{Additional Quantitative Analysis}
\label{appx:quantitative}

\begin{table*}[!htb] %
    \centering
    \setlength{\tabcolsep}{3pt}
    \caption{\textbf{Quantitative Comparison of Multi-Concept Compositions.} We evaluate different methods using CLIP and DINO similarity scores on instance segmentation maps. Our method consistently outperforms others across all metrics, achieving the highest minimum, average, and maximum similarity scores.} \label{tab:quantitative_comparison}
    \begin{tabular}{c c | c | c | c | c | c}
        \toprule
        & & Merge & Composite  & ZipLoRA & Mix-of-Show & \textbf{Ours} \\ \midrule
        \multirow{3}{*}{\rotatebox[origin=c]{90}{\textbf{CLIP}}} 
        & Min. & 76.0\% $\pm$ 8.7\% & 76.2\% $\pm$ 7.2\% & 73.4\%  $\pm$ 8.1\% & 75.2\%  $\pm$ 9.5\% & \textbf{83.3\% $\pm$ 5.5\%} \\
        & Avg. & 79.5\% $\pm$ 8.3\%  & 79.7\% $\pm$ 6.8\%  & 77.1\% $\pm$ 7.6\%  & 78.7\% $\pm$ 9.2\%  & \textbf{87.1\% $\pm$ 4.9\% } \\
        & Max. & 82.5\% $\pm$ 8.1\%  & 82.5\% $\pm$ 6.7\%  & 80.6\% $\pm$ 7.6\%  & 81.7\% $\pm$ 9.2\%  & \textbf{89.8\% $\pm$ 4.8\%}  \\ \midrule
        \multirow{3}{*}{\rotatebox[origin=c]{90}{\textbf{DINO}}} 
        & Min. & 37.0\% $\pm$ 15\% & 30.3\% $\pm$ 13\% & 36.9\%  $\pm$ 13\% & 37.5\% $\pm$ 17\% & \textbf{47.2\% $\pm$ 14\%} \\
        & Avg. & 43.7\% $\pm$ 17\% & 38.5\% $\pm$ 13\% & 49.6\% $\pm$ 15\% & 48.0\% $\pm$ 22\% & \textbf{57.3\% $\pm$ 14\%} \\
        & Max. & 50.5\% $\pm$ 17\% & 49.5\% $\pm$ 14\% & 53.3\% $\pm$ 16\% & 55.6\% $\pm$ 23\% & \textbf{69.1\% $\pm$ 14\%} \\ 
        \bottomrule
        
    \end{tabular}

\end{table*}

In addition to the results presented in the main paper, we apply further experiments to assess the performance of our method in detail. Specifically, we apply instance segmentation methods to the composed images to identify and isolate object instances. For this, we use SEEM~\cite{zou2024segment} to segment the objects within the images. After segmentation, we calculate the similarity metrics separately for each object instance, allowing for a more granular comparison of the methods. We perform these evaluations on a set of 700 images per method, as shown in \cref{tab:quantitative_comparison}. The results demonstrate that our method significantly outperforms others across multiple metrics. In particular, we calculate DINO scores, which further highlight the effectiveness of our approach compared to competing methods. Moreover, we also compute CLIP scores as additional evidence of our method's superior performance.

\section{Comparison with Orthogonal Adaptation}
\begin{figure}[!htb]
    \centering
    \includegraphics[width=\linewidth]{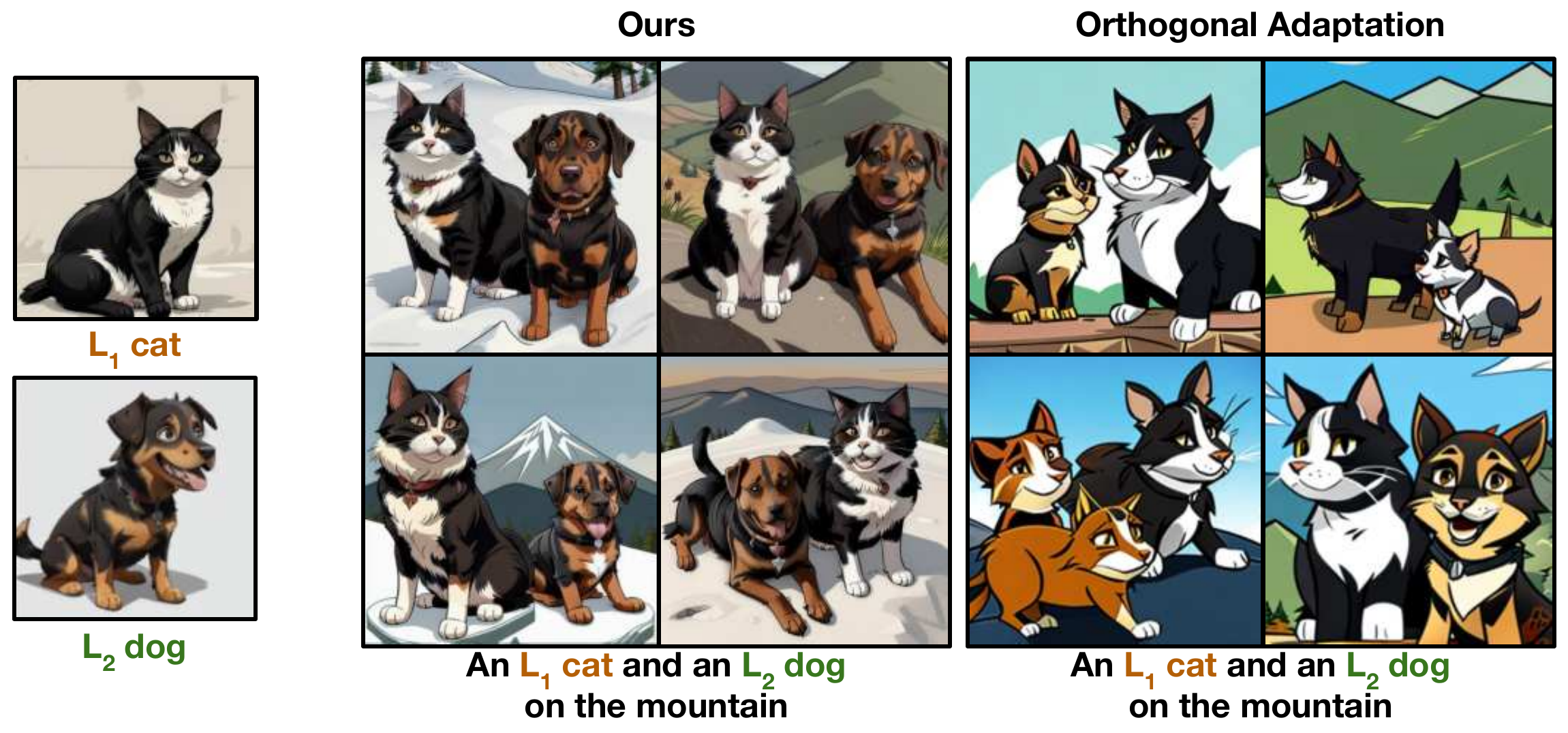}
    \caption{{\textbf{Qualitative comparison with Orthogonal Adaptation.} \methodName{} achieves consistent multi-concept compositions without additional conditions like sketches or key points, unlike Orthogonal Adaptation. Without user-provided conditions, Orthogonal Adaptation method fails to generate the accurate depictions.}}\label{fig:orthogonal_comparison}
\end{figure}

We compare \methodName{} with Orthogonal Adaptation, which operates by enforcing constraints to separate attributes across LoRAs. While this method reduces interference, it requires additional user-provided conditions, such as sketches or key points, to ensure accurate multi-concept compositions.
In contrast, \methodName{} achieves consistent multi-concept compositions without relying on extra conditions. As shown in Figure~\ref{fig:orthogonal_comparison}, Orthogonal Adaptation struggles to generate accurate depictions when such conditions are absent. Our approach seamlessly integrates multiple LoRA modelss while preserving individual attributes, leading to more coherent and natural compositions.

\section{Additional Qualitative Results}
\label{appx:comparison}
\begin{figure*}[!htbp]
    \centering
    \includegraphics[width=\linewidth]{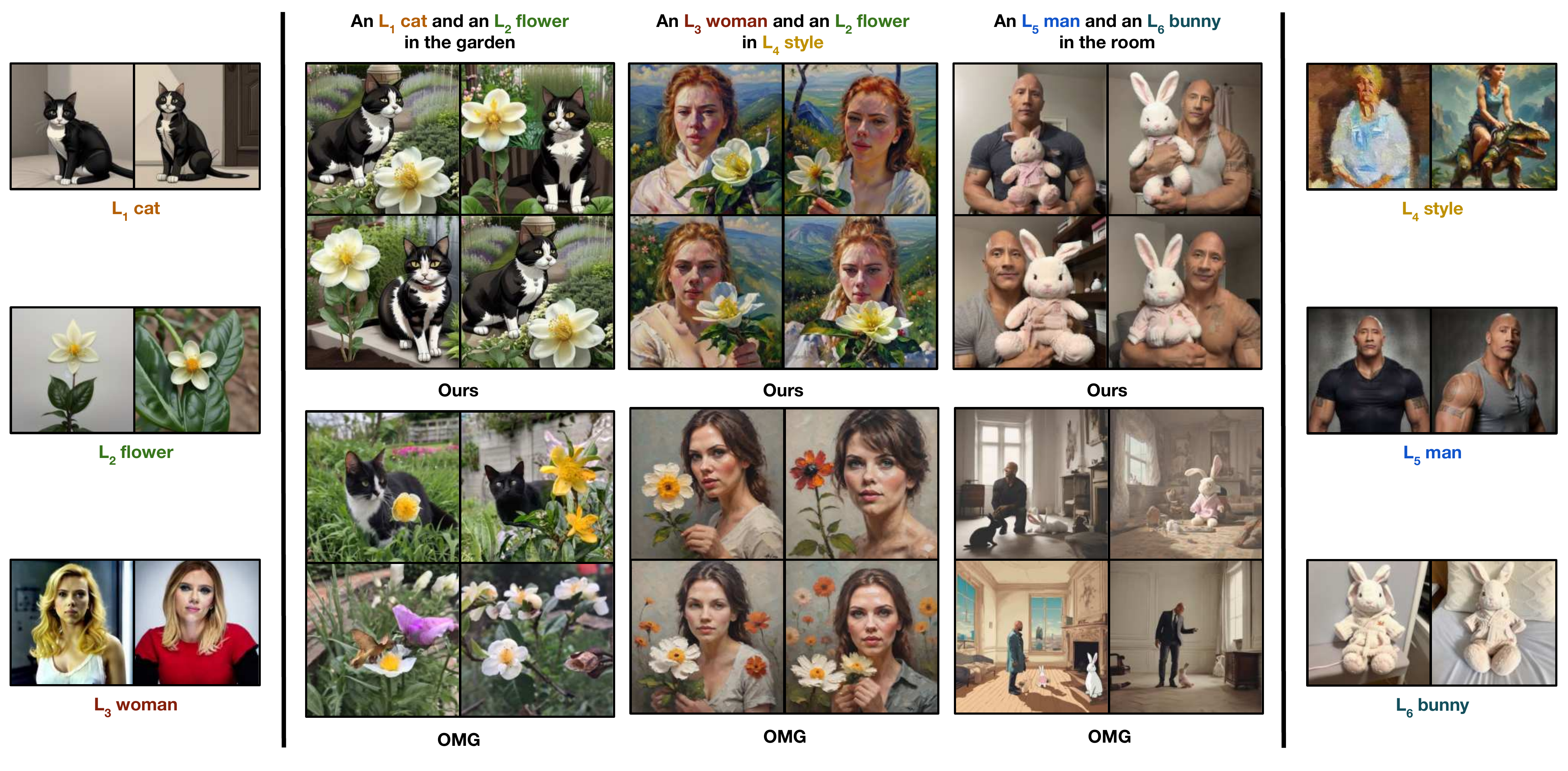}
    \caption{\textbf{Qualitative comparison with OMG.} Our method (top row) consistently produces more coherent and accurate compositions compared to OMG (bottom row). By leveraging attention map updates and latent fusion, \methodName{} effectively handles multi-concept generation without relying on segmentation, leading to higher quality results, particularly in complex scenes.}
    \label{fig:omg_comparison}
\end{figure*}

\paragraph{Comparison with OMG.}
We perform a qualitative comparison between our method, \methodName{}, and OMG~\cite{kong2024omg}. OMG relies on off-the-shelf segmentation methods to isolate subjects before generating images. As seen in Fig.~\ref{fig:omg_comparison},   while this enables well-defined subject boundaries, the performance of OMG is heavily dependent on the accuracy of the segmentation model. Errors in segmentation can result in incomplete or incorrect generation, particularly in complex scenes involving multiple interacting subject. For instance, if the segmentation model fails to detect a flower, this may prevent the correct placement of the LoRA in the composition (see Fig. \ref{fig:omg_comparison} bottom-left).  Moreover, since OMG depends on the base image generated by the Stable Diffusion model, it also encounters the attention overlap and attribute binding issues identified by \cite{chefer2023attend}. For instance, if the Stable Diffusion model does not generate the required objects in the base image from the text prompt 'A man and a bunny in the room', then OMG cannot produce the desired composition. This issue is apparent in Fig. \ref{fig:omg_comparison}, where the rightmost image shows that the base model generated only a bunny, omitting the man.  In contrast, \methodName{} bypasses the need for explicit segmentation by directly updating attention maps and fusing latent representations. This ensures that each concept, represented by different LoRA models, is accurately captured and preserved during generation. The comparison in Fig.~\ref{fig:omg_comparison} demonstrates that \methodName{} produces more coherent compositions, maintaining the integrity of each concept even in challenging multi-concept scenarios.

\begin{figure*}[!htbp]
  \centering
  \includegraphics[width=\linewidth]{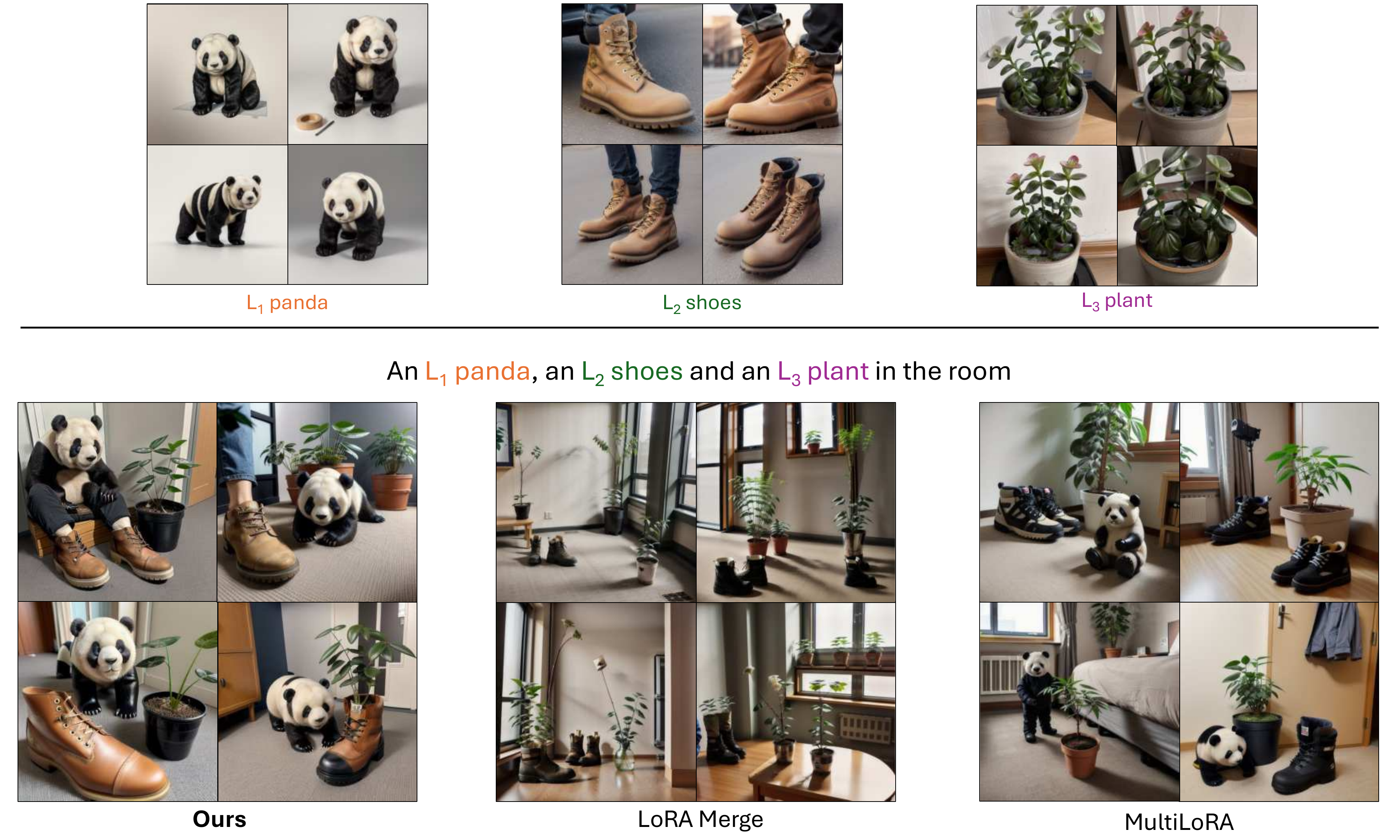}
  \caption{\textbf{Qualitative comparison of \methodName{}} with other LoRA methods using 3 LoRAs to generate a single image. Our approach consistently produces images that more accurately reflect the input text prompts, LoRA subjects, and LoRA styles.}
  \label{fig:3loras2}
\end{figure*}

\paragraph{Extensive Qualitative Results.} The rest of the Supplementary Materials will provide additional qualitative comparisons which contain the following competitors: Mix-of-Show~\cite{gu2023mix}, MultiLoRA~\cite{zhong2024multi}, LoRA-Merge~\cite{ryu2023low}, ZipLoRA~\cite{shah2023ziplora}, and Custom Diffusion~\cite{kumari2023multi} on various LoRAs and prompts. Figure \ref{fig:3loras2} compare LoRA-Merge and MultiLoRA using three combined LoRAs, while later figures expand the comparison to include all methods across two separate LoRAs.

\begin{figure*}[!htbp]
  \centering
  \includegraphics[width=\linewidth]{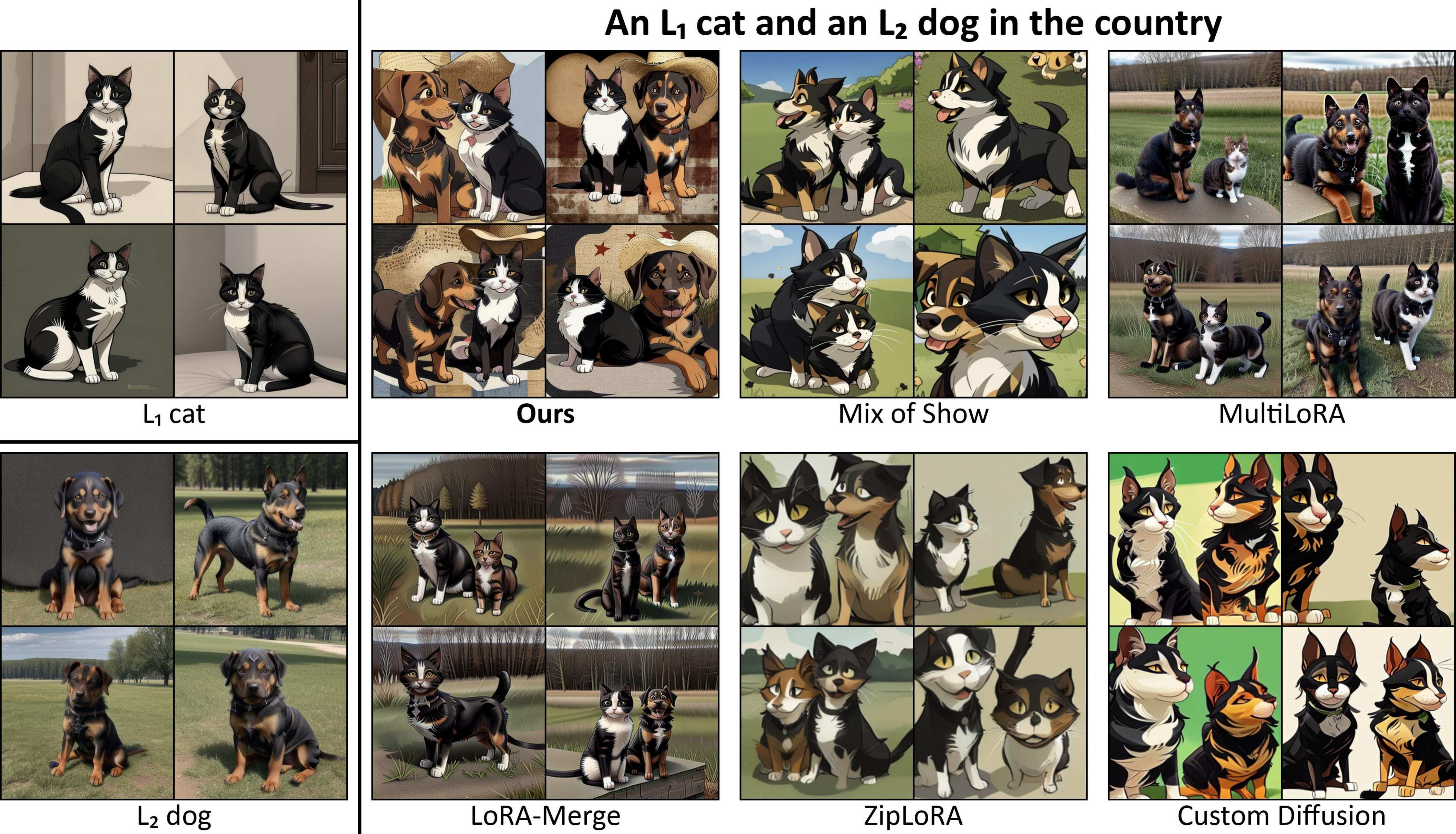}
  \caption{\textbf{Qualitative comparison of \methodName{}} with other LoRA methods. Our approach consistently produces images that more accurately reflect the input text prompts, LoRA subjects, and LoRA styles.}
  \label{fig:results1}
\end{figure*}

\begin{figure*}[!htbp]
  \centering
  \includegraphics[width=\linewidth]{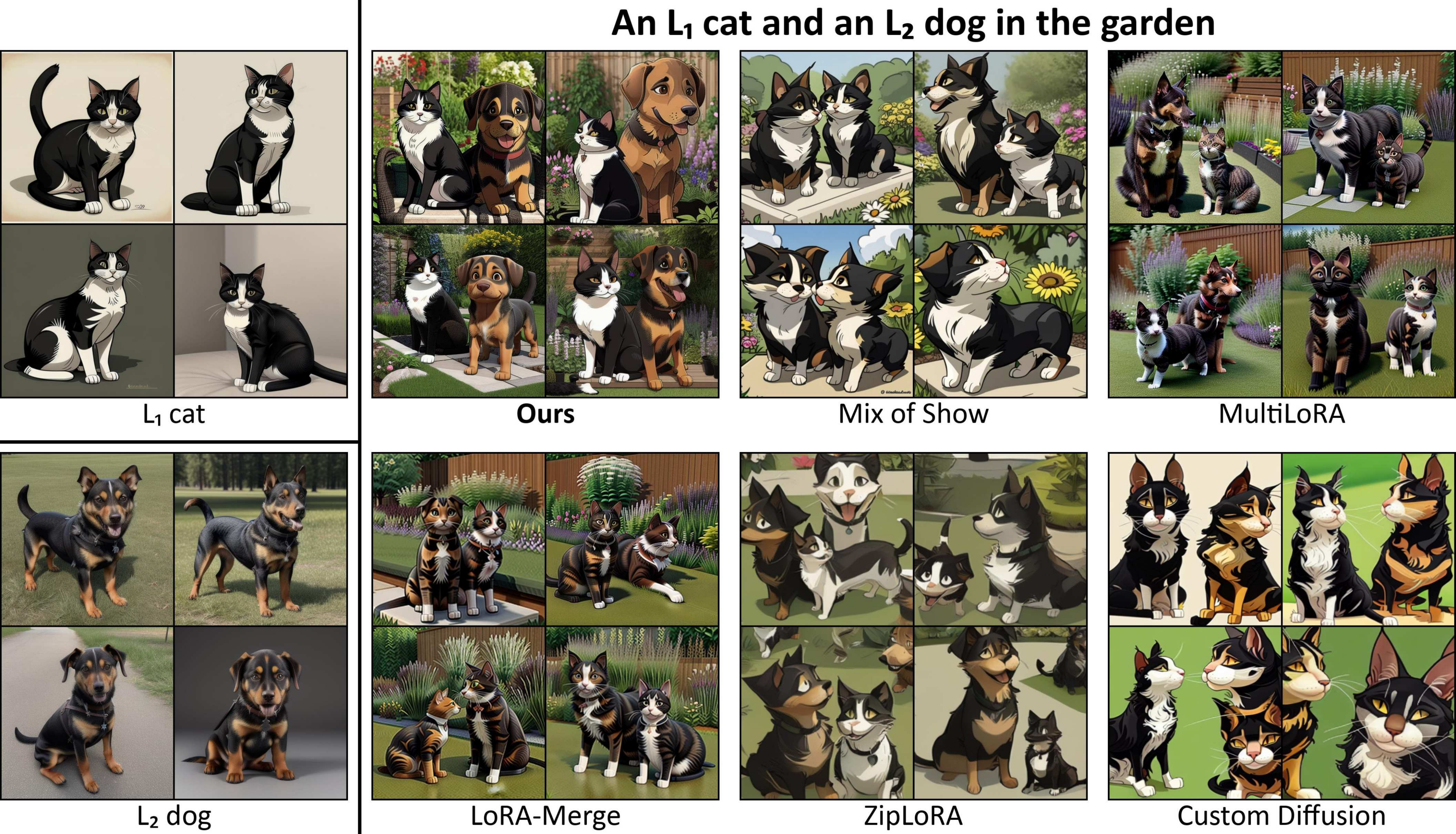}
  \caption{\textbf{Qualitative comparison of \methodName{}} with other LoRA methods. Our approach consistently produces images that more accurately reflect the input text prompts, LoRA subjects, and LoRA styles.}
  \label{fig:results2}
\end{figure*}

\begin{figure*}[!htbp]
  \centering
  \includegraphics[width=\linewidth]{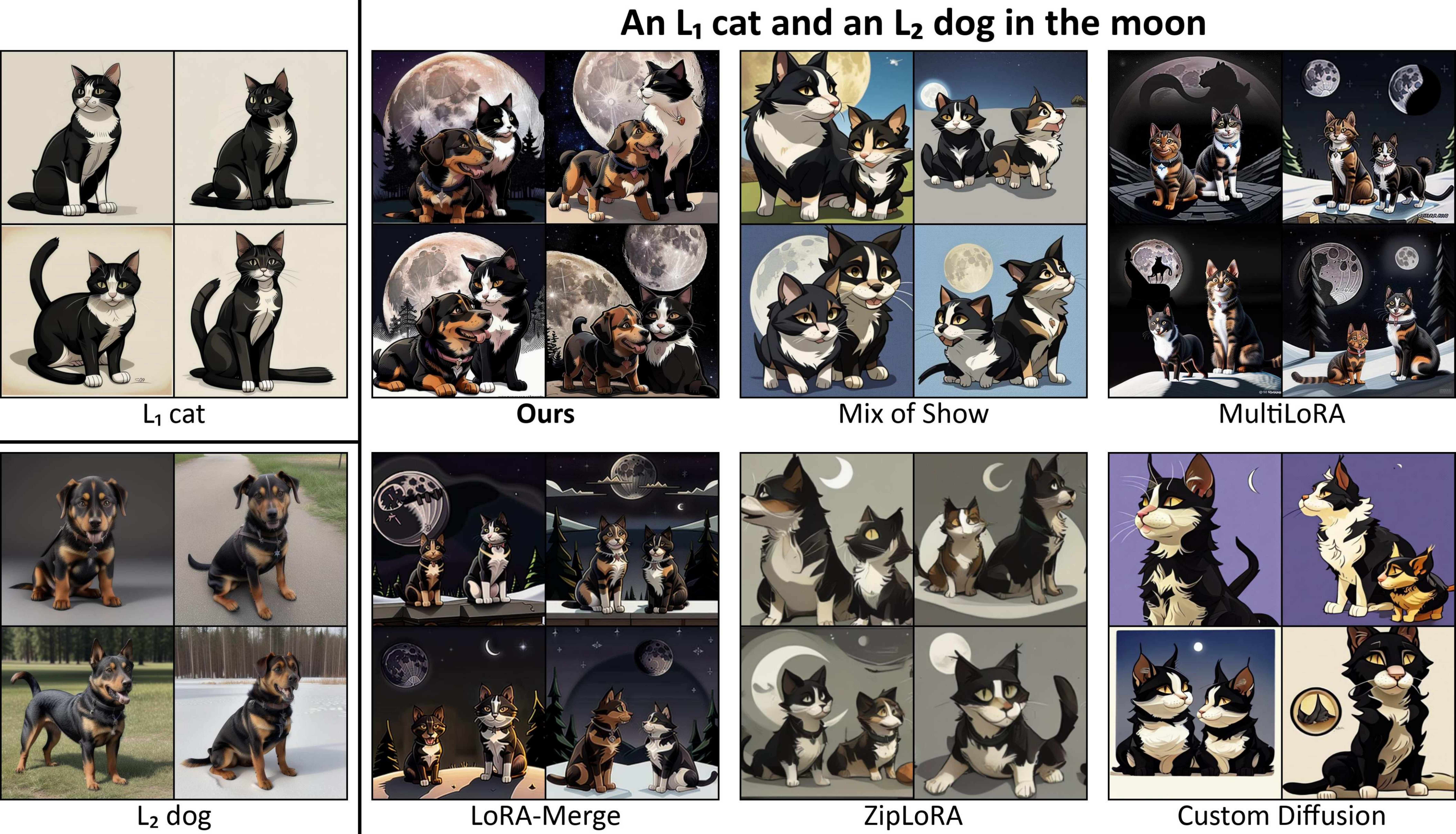}
  \caption{\textbf{Qualitative comparison of \methodName{}} with other LoRA methods. Our approach consistently produces images that more accurately reflect the input text prompts, LoRA subjects, and LoRA styles.}
  \label{fig:results3}
\end{figure*}

\begin{figure*}[!htbp]
  \centering
  \includegraphics[width=\linewidth]{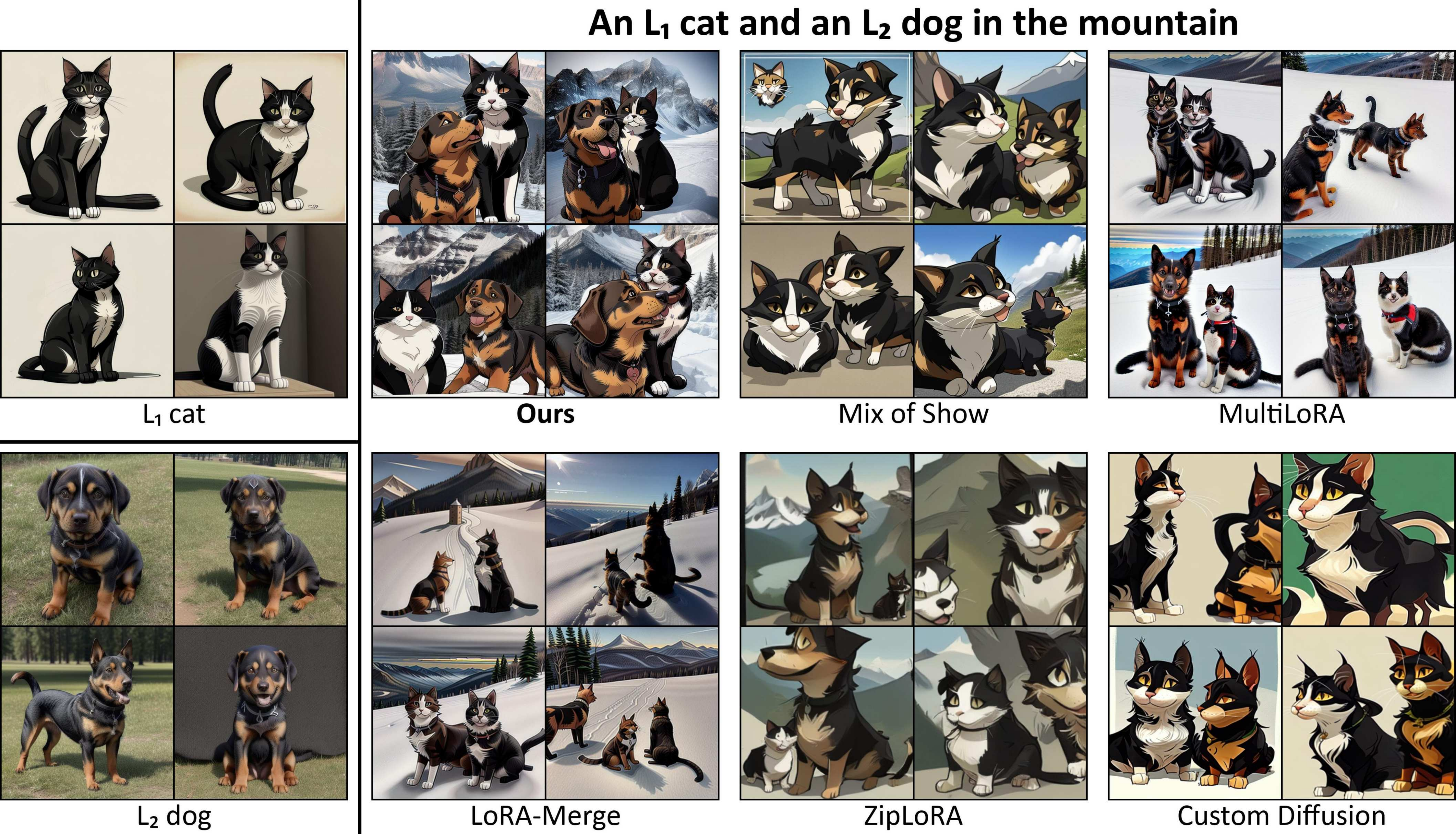}
  \caption{\textbf{Qualitative comparison of \methodName{}} with other LoRA methods. Our approach consistently produces images that more accurately reflect the input text prompts, LoRA subjects, and LoRA styles.}
  \label{fig:results4}
\end{figure*}

\begin{figure*}[!htbp]
  \centering
  \includegraphics[width=\linewidth]{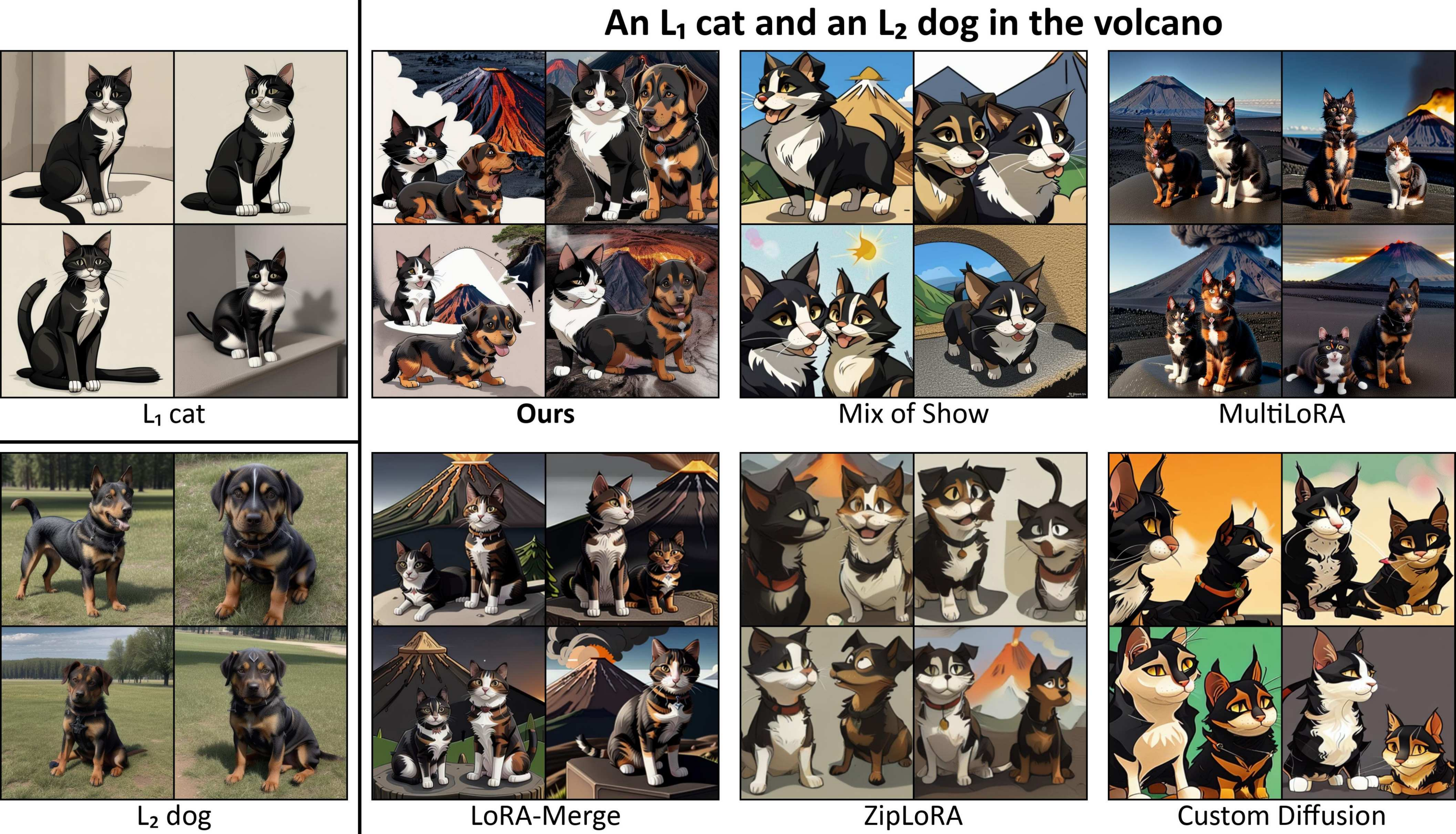}
  \caption{\textbf{Qualitative comparison of \methodName{}} with other LoRA methods. Our approach consistently produces images that more accurately reflect the input text prompts, LoRA subjects, and LoRA styles.}
  \label{fig:results5}
\end{figure*}

\begin{figure*}[!htbp]
  \centering
  \includegraphics[width=\linewidth]{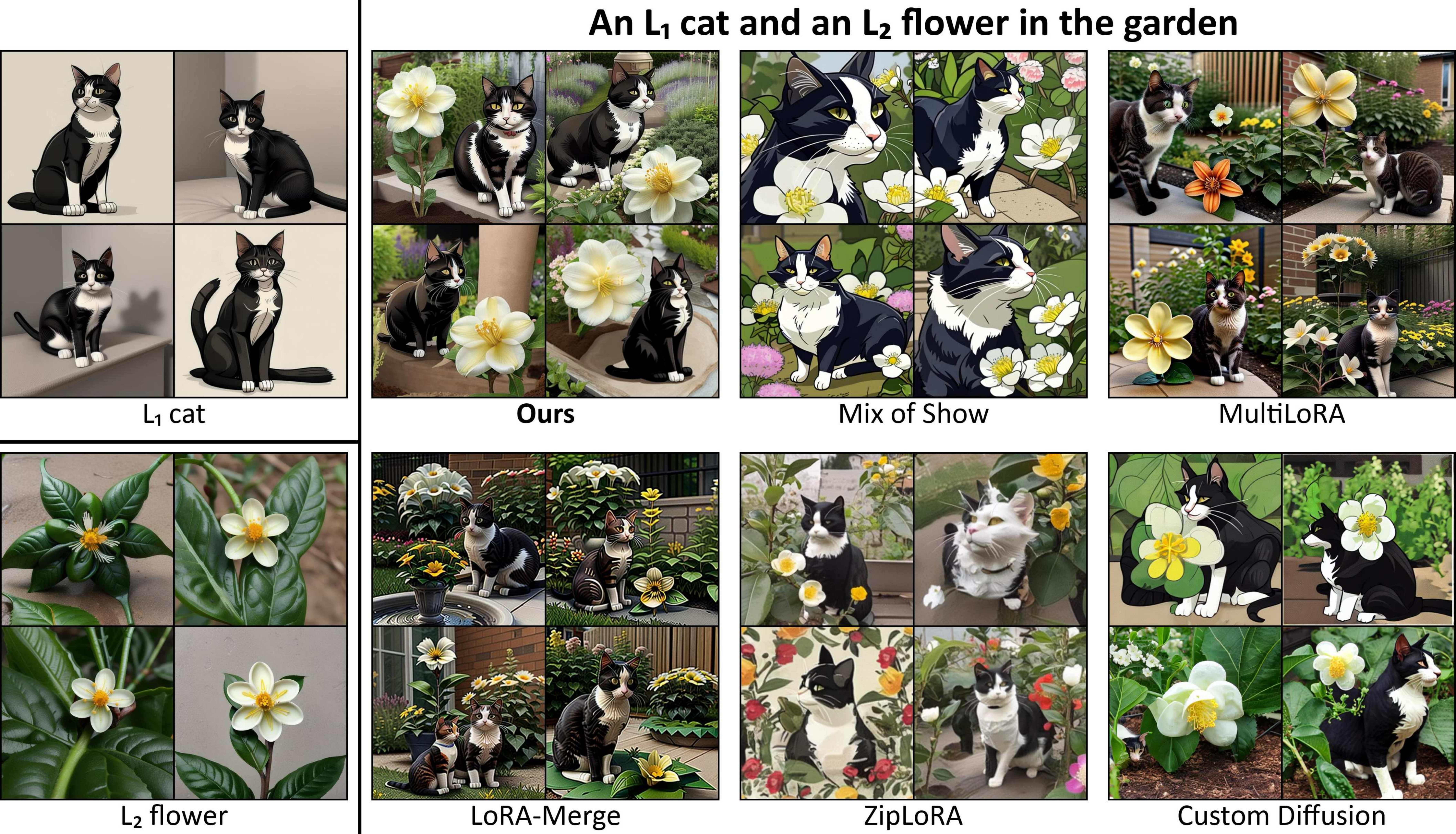}
  \caption{\textbf{Qualitative comparison of \methodName{}} with other LoRA methods. Our approach consistently produces images that more accurately reflect the input text prompts, LoRA subjects, and LoRA styles.}
  \label{fig:results6}
\end{figure*}

\begin{figure*}[!htbp]
  \centering
  \includegraphics[width=\linewidth]{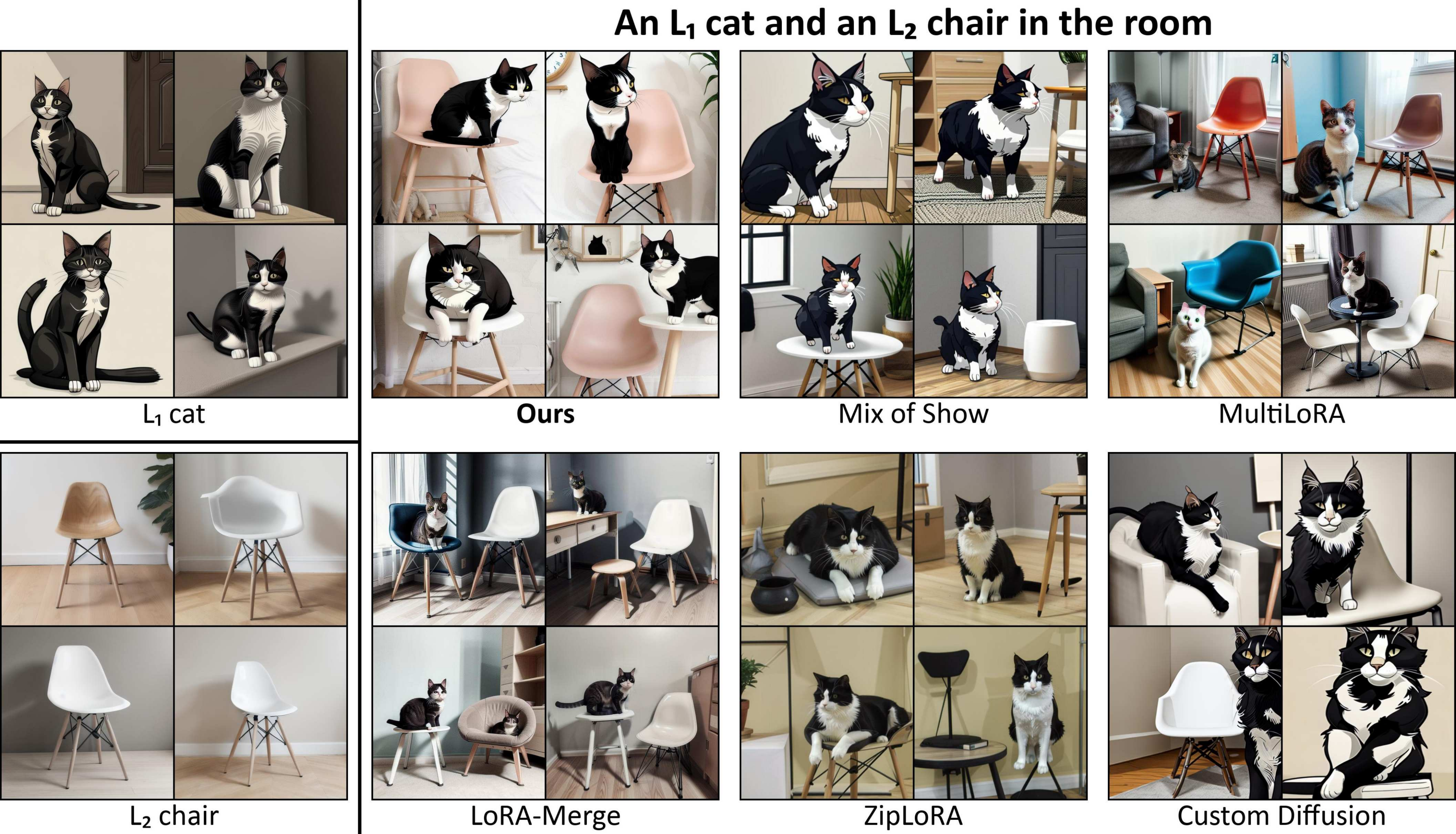}
  \caption{\textbf{Qualitative comparison of \methodName{}} with other LoRA methods. Our approach consistently produces images that more accurately reflect the input text prompts, LoRA subjects, and LoRA styles.}
  \label{fig:results7}
\end{figure*}

\begin{figure*}[!htbp]
  \centering
  \includegraphics[width=\linewidth]{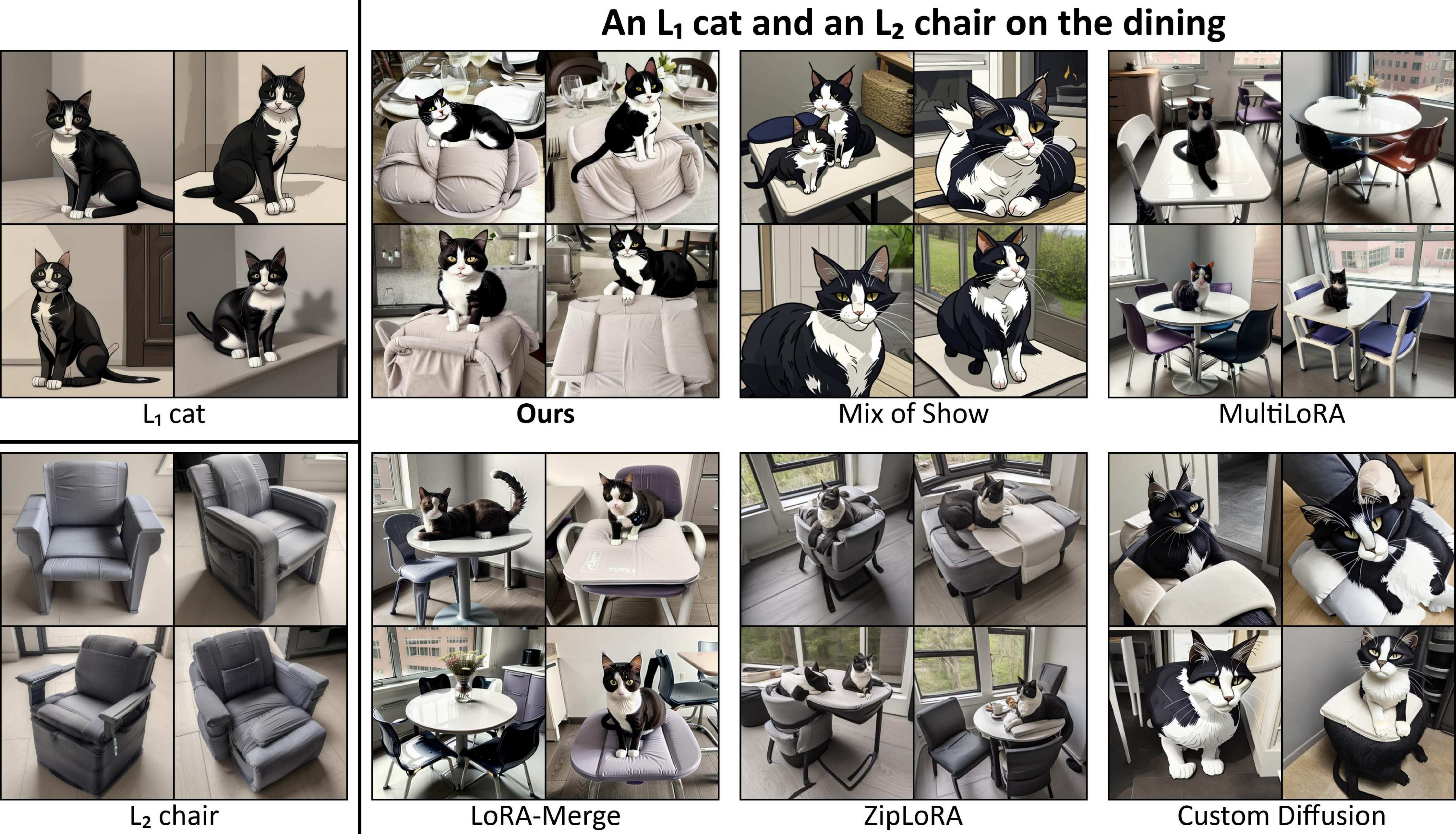}
  \caption{\textbf{Qualitative comparison of \methodName{}} with other LoRA methods. Our approach consistently produces images that more accurately reflect the input text prompts, LoRA subjects, and LoRA styles.}
  \label{fig:results8}
\end{figure*}

\begin{figure*}[!htbp]
  \centering
  \includegraphics[width=\linewidth]{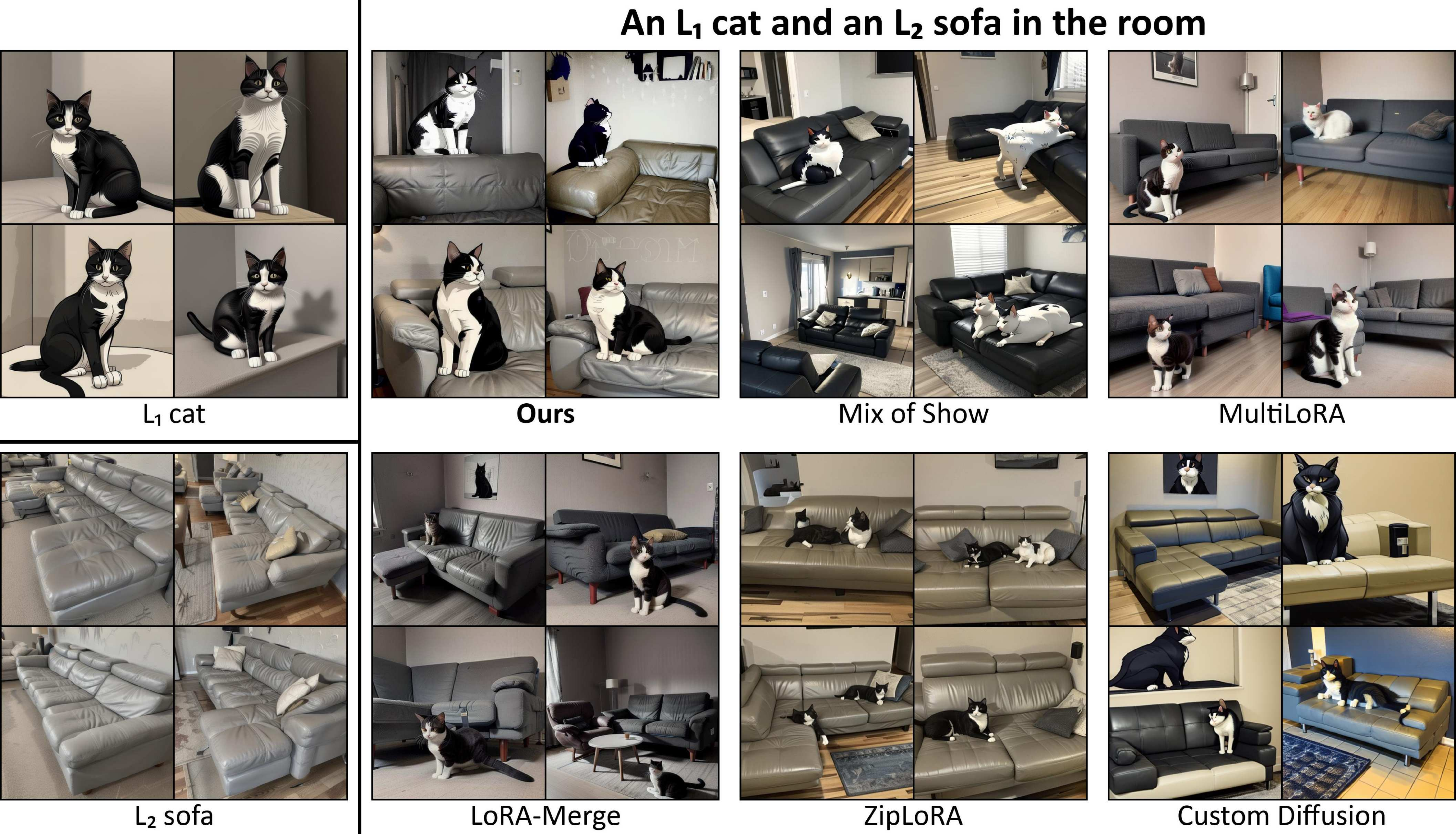}
  \caption{\textbf{Qualitative comparison of \methodName{}} with other LoRA methods. Our approach consistently produces images that more accurately reflect the input text prompts, LoRA subjects, and LoRA styles.}
  \label{fig:results9}
\end{figure*}

\begin{figure*}[!htbp]
  \centering
  \includegraphics[width=\linewidth]{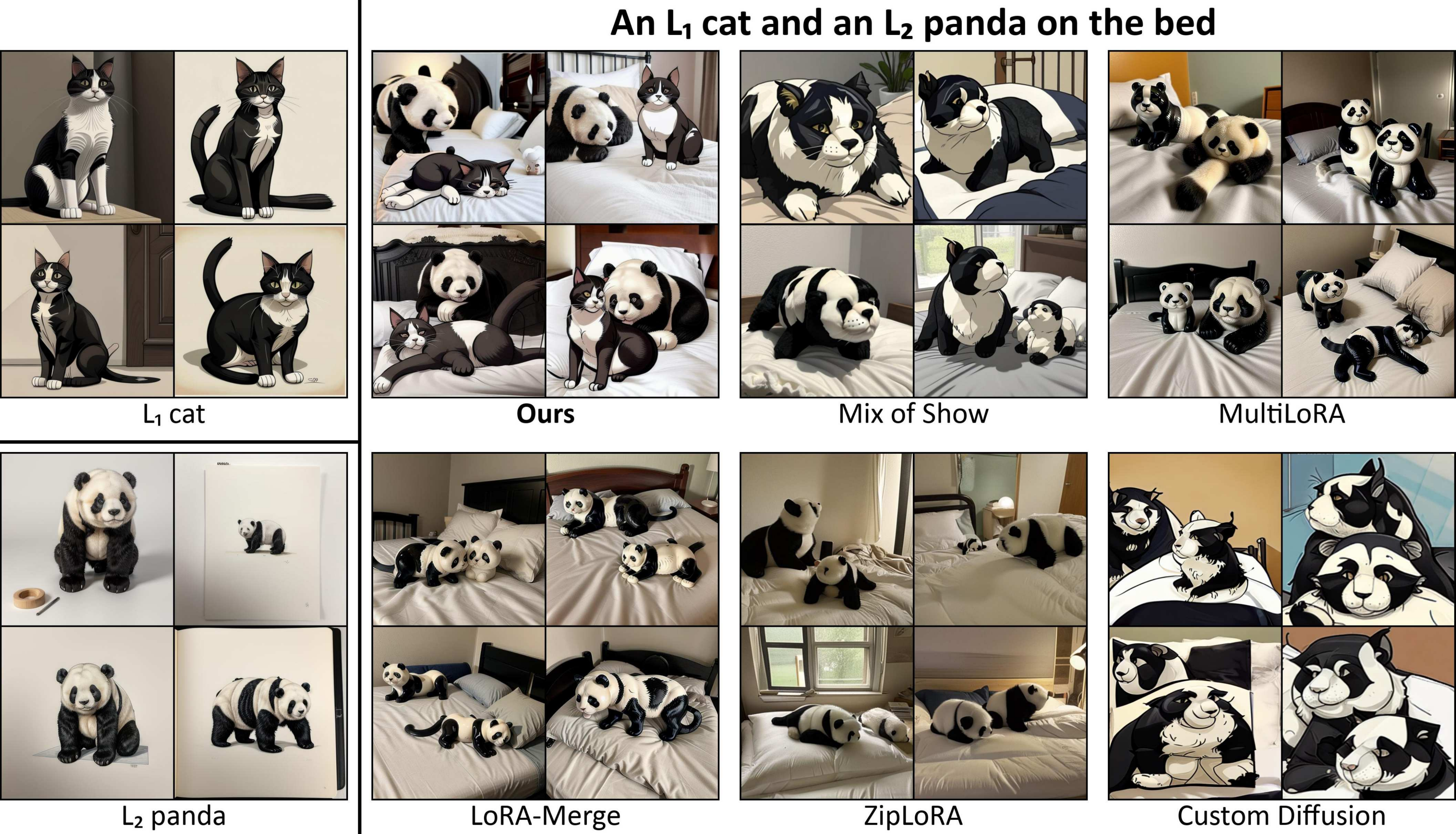}
  \caption{\textbf{Qualitative comparison of \methodName{}} with other LoRA methods. Our approach consistently produces images that more accurately reflect the input text prompts, LoRA subjects, and LoRA styles.}
  \label{fig:results10}
\end{figure*}

\begin{figure*}[!htbp]
  \centering
  \includegraphics[width=\linewidth]{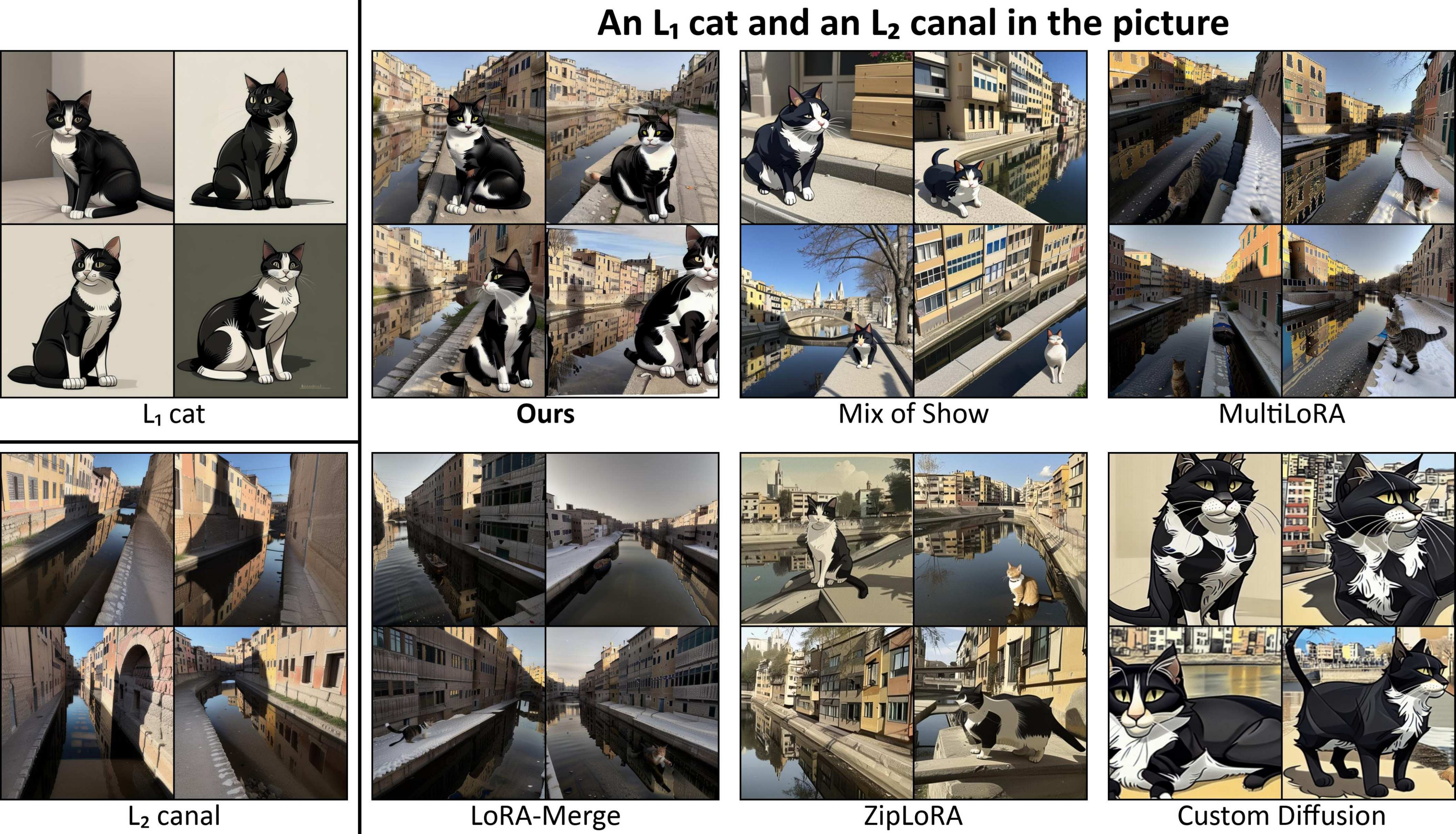}
  \caption{\textbf{Qualitative comparison of \methodName{}} with other LoRA methods. Our approach consistently produces images that more accurately reflect the input text prompts, LoRA subjects, and LoRA styles.}
  \label{fig:results11}
\end{figure*}

\begin{figure*}[!htbp]
  \centering
  \includegraphics[width=\linewidth]{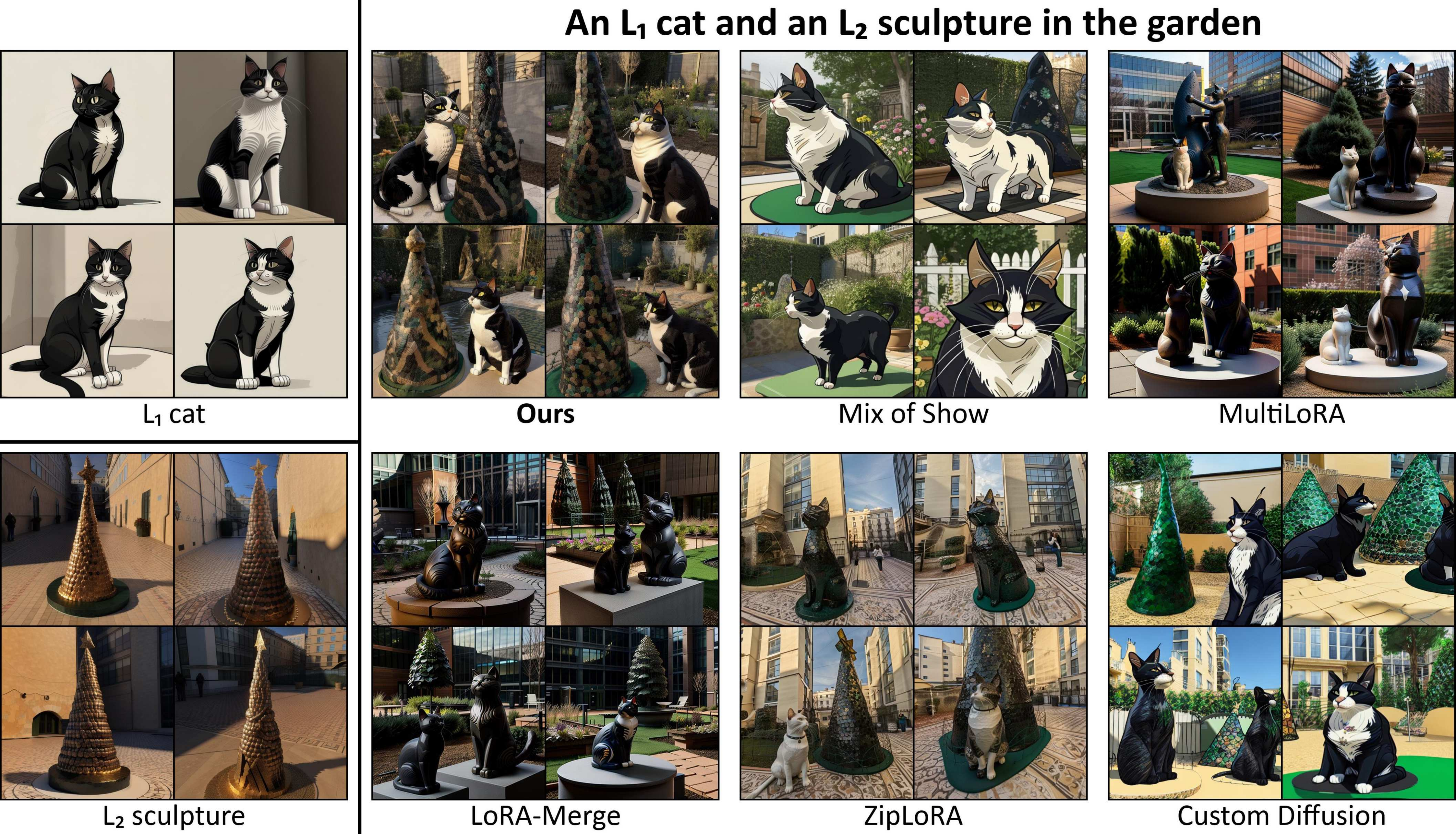}
  \caption{\textbf{Qualitative comparison of \methodName{}} with other LoRA methods. Our approach consistently produces images that more accurately reflect the input text prompts, LoRA subjects, and LoRA styles.}
  \label{fig:results12}
\end{figure*}

\begin{figure*}[!htbp]
  \centering
  \includegraphics[width=\linewidth]{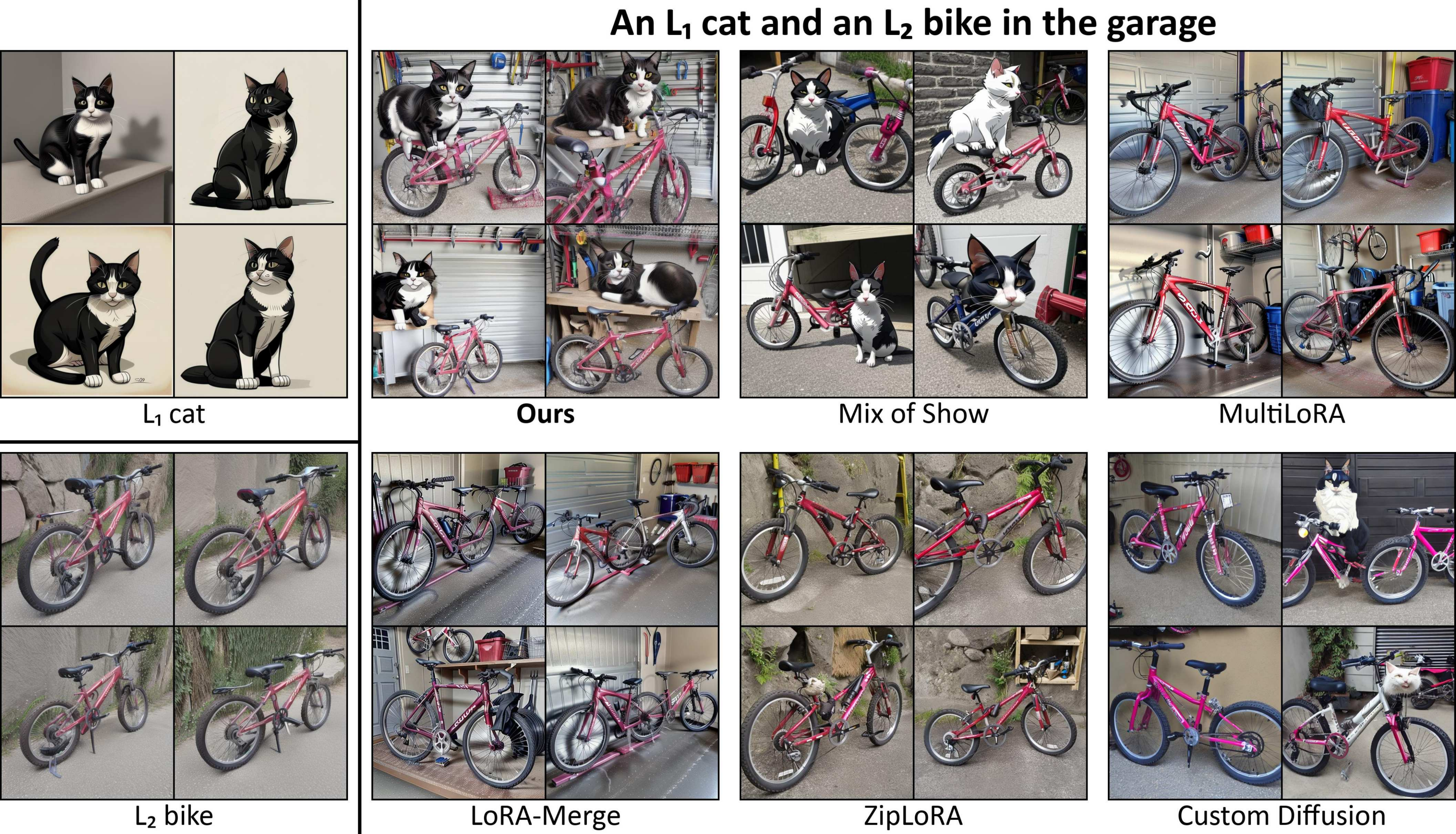}
  \caption{\textbf{Qualitative comparison of \methodName{}} with other LoRA methods. Our approach consistently produces images that more accurately reflect the input text prompts, LoRA subjects, and LoRA styles.}
  \label{fig:results13}
\end{figure*}

\begin{figure*}[!htbp]
  \centering
  \includegraphics[width=\linewidth]{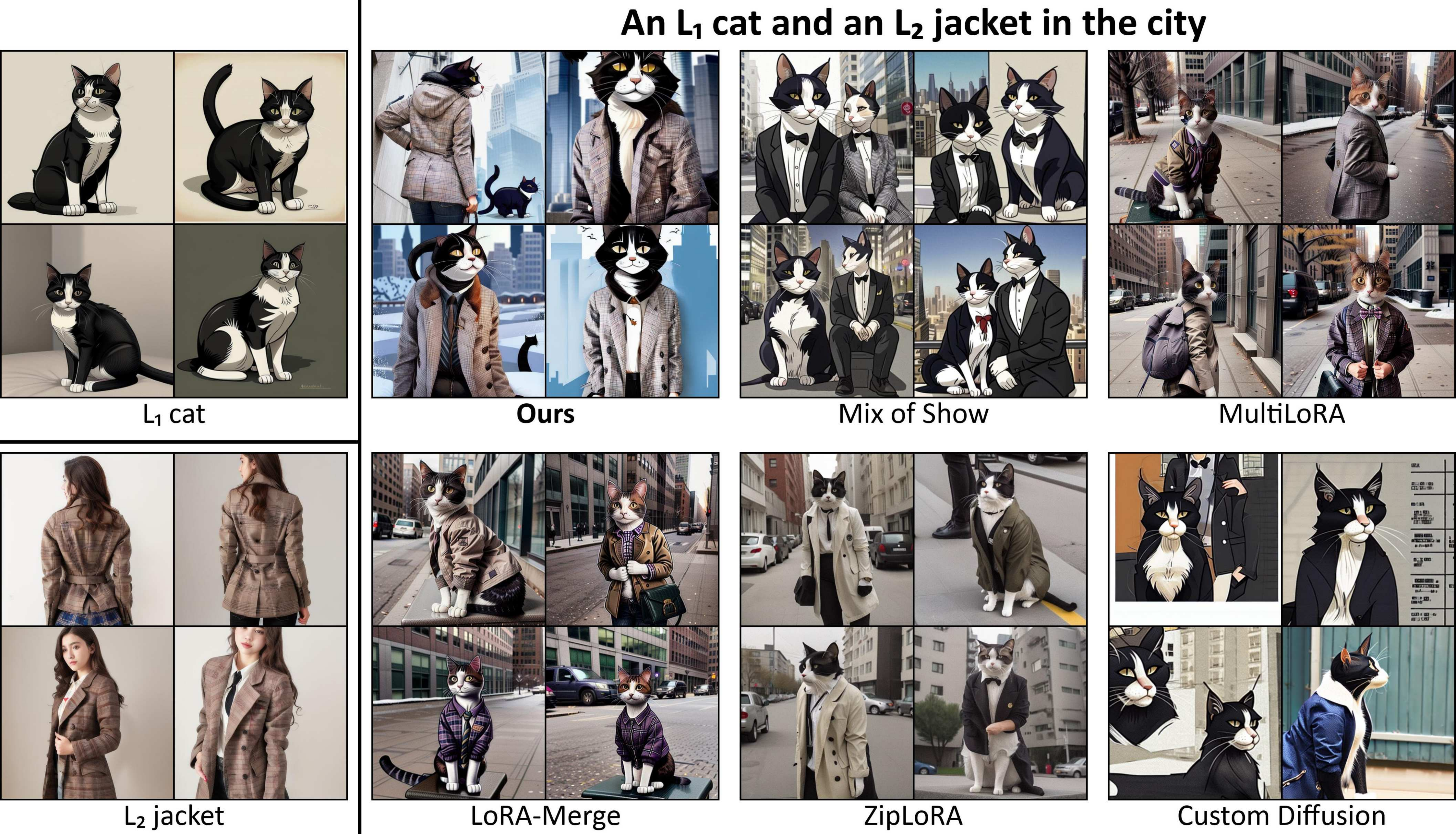}
  \caption{\textbf{Qualitative comparison of \methodName{}} with other LoRA methods. Our approach consistently produces images that more accurately reflect the input text prompts, LoRA subjects, and LoRA styles.}
  \label{fig:results14}
\end{figure*}

\begin{figure*}[!htbp]
  \centering
  \includegraphics[width=\linewidth]{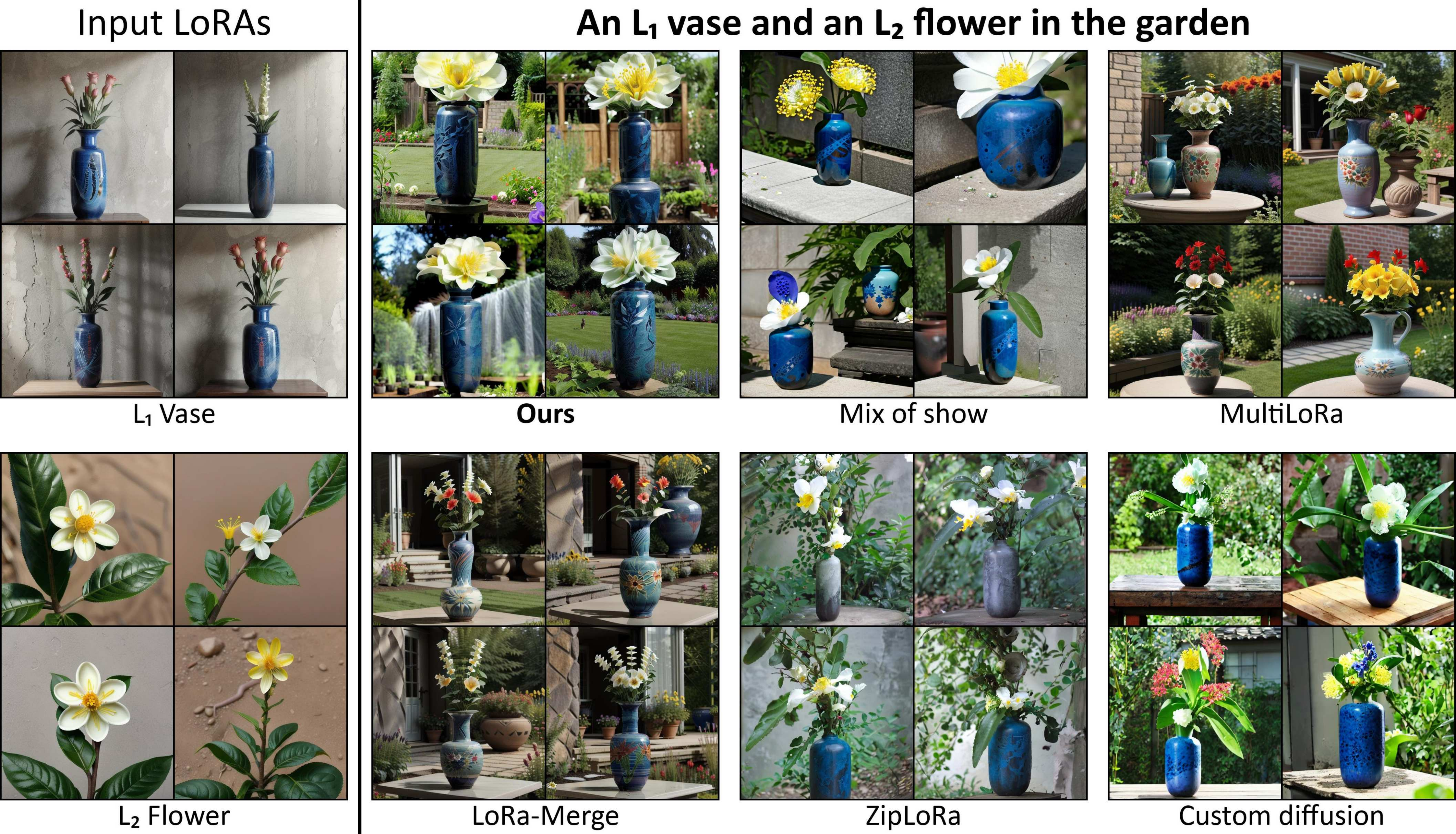}
  \caption{\textbf{Qualitative comparison of \methodName{}} with other LoRA methods. Our approach consistently produces images that more accurately reflect the input text prompts, LoRA subjects, and LoRA styles.}
  \label{fig:results15}
\end{figure*}

\begin{figure*}[!htbp]
  \centering
  \includegraphics[width=\linewidth]{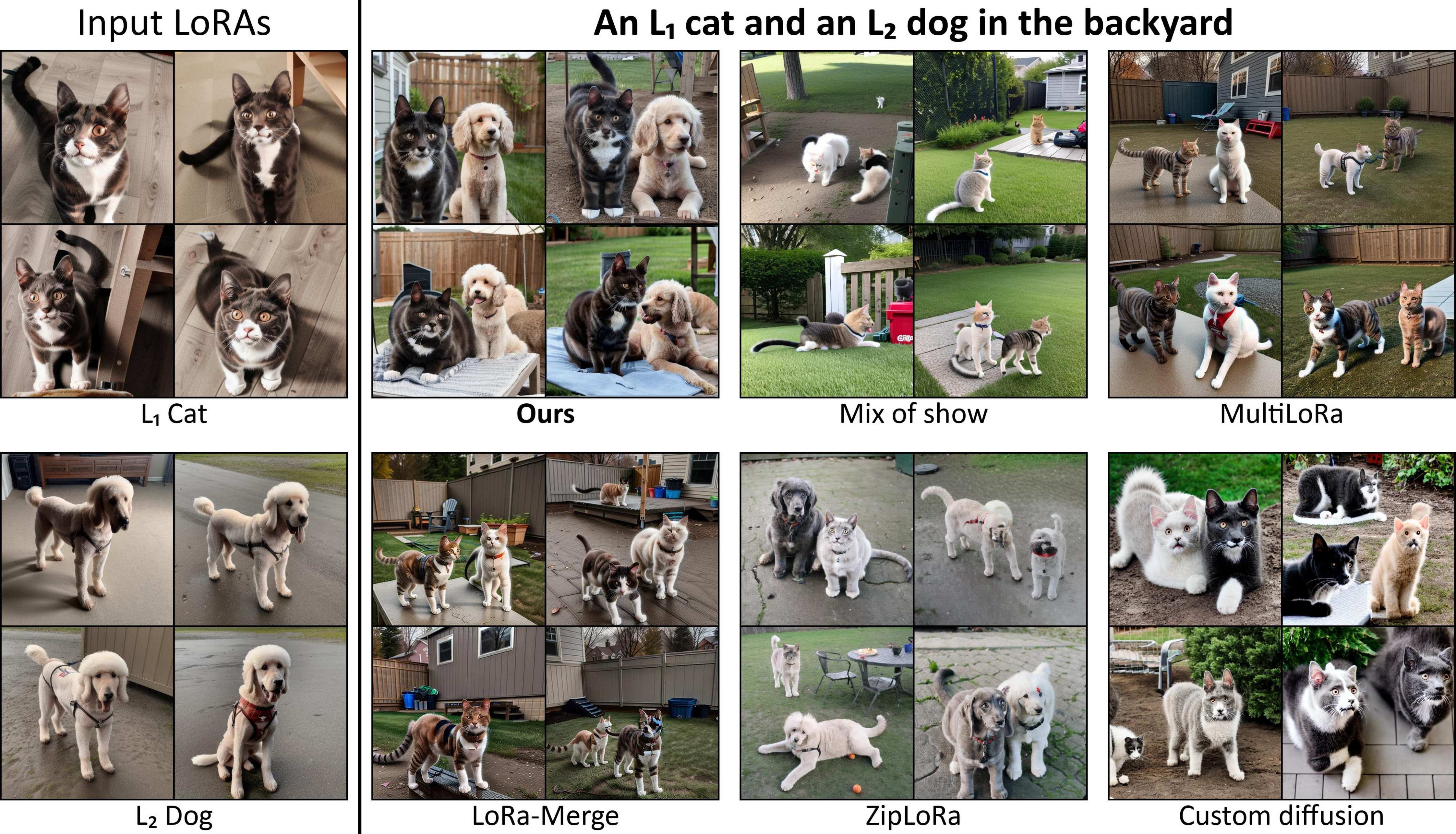}
  \caption{\textbf{Qualitative comparison of \methodName{}} with other LoRA methods. Our approach consistently produces images that more accurately reflect the input text prompts, LoRA subjects, and LoRA styles.}
  \label{fig:results18}
\end{figure*}

\begin{figure*}[!htbp]
  \centering
  \includegraphics[width=\linewidth]{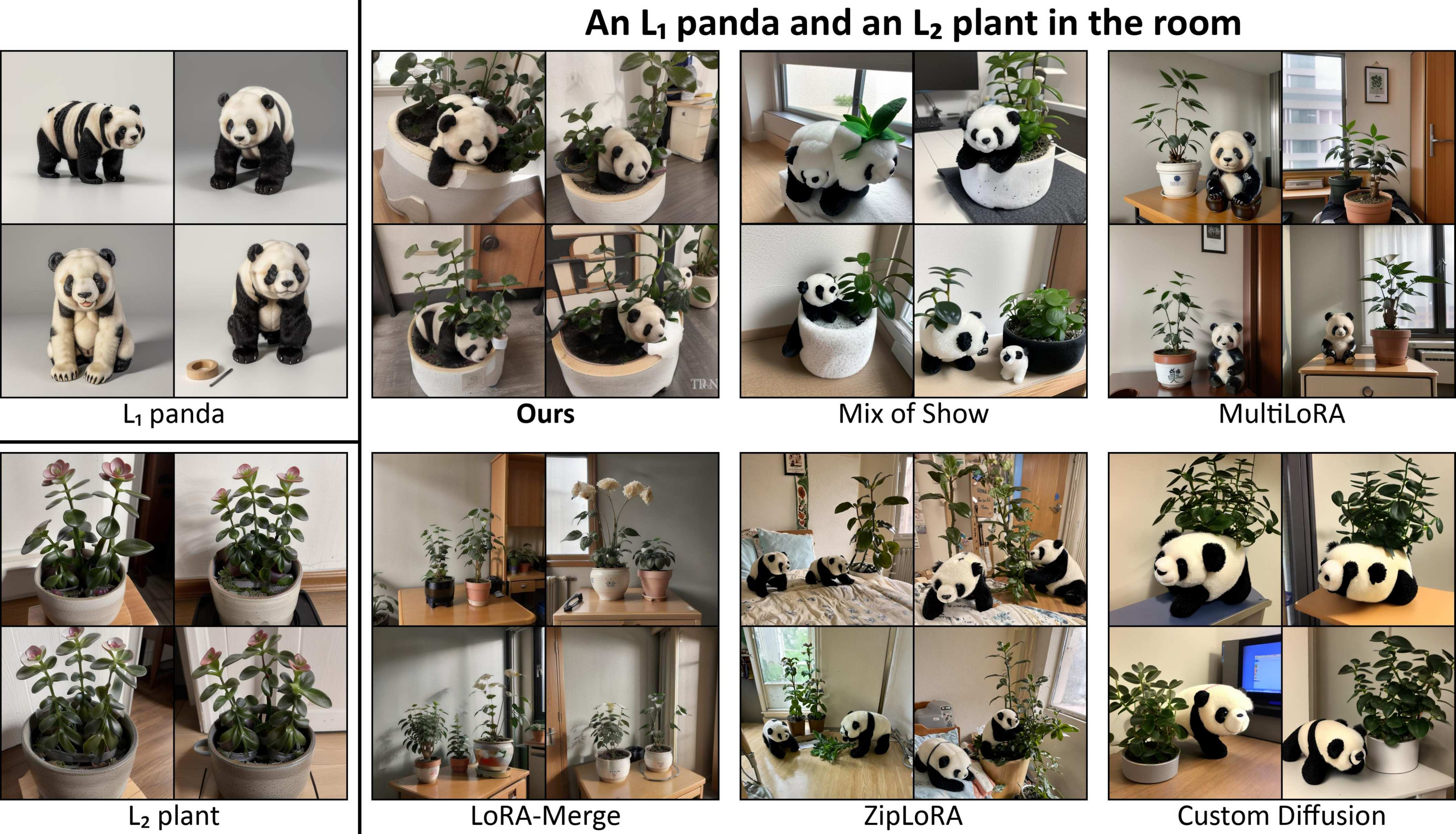}
  \caption{\textbf{Qualitative comparison of \methodName{}} with other LoRA methods. Our approach consistently produces images that more accurately reflect the input text prompts, LoRA subjects, and LoRA styles.}
  \label{fig:results20}
\end{figure*}

\begin{figure*}[!htbp]
  \centering
  \includegraphics[width=\linewidth]{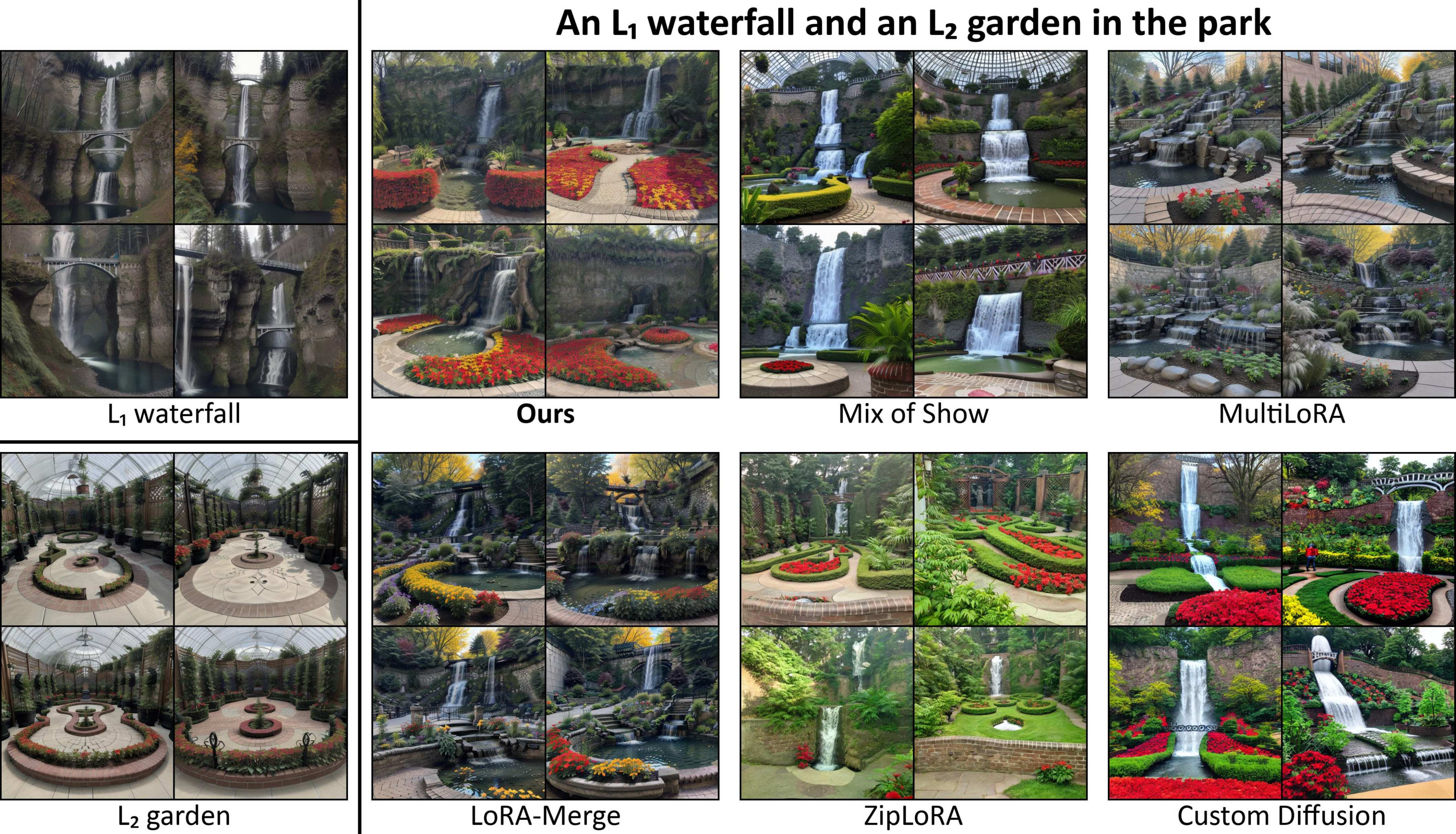}
  \caption{\textbf{Qualitative comparison of \methodName{}} with other LoRA methods. Our approach consistently produces images that more accurately reflect the input text prompts, LoRA subjects, and LoRA styles.}
  \label{fig:results21}
\end{figure*}